\documentclass[journal,10pt]{IEEEtran}

\usepackage{ifpdf}
\ifpdf
\usepackage[pdftex]{graphicx}
\usepackage{epstopdf}

\else
\usepackage{graphicx}
\usepackage{epsfig,psfrag}
\psdraft\psfull
\fi

\usepackage[noend]{algpseudocode}
\usepackage{algorithm}
\usepackage{graphicx}
\usepackage{float}
\usepackage{fourier}
\usepackage{subfigure}
\usepackage{setspace}
\usepackage{color}
\usepackage{cite}
\usepackage{comment} 
\usepackage{amsmath}

\usepackage{color,subfigure}
\usepackage{amsmath,amssymb,amsfonts}
\usepackage{url}
\usepackage{setspace}
\usepackage{enumerate}
\usepackage{flushend}
\usepackage{fancyhdr}
\usepackage{multirow}
\usepackage{footnote}
\makesavenoteenv{tabular}

\renewcommand{\baselinestretch}{0.93}

\title{\LARGE \bf
Real Time Motion Planning Using Constrained Iterative Linear Quadratic Regulator for On-Road Self-Driving
}

\makeatletter          
\let\@fnsymbol\@arabic
\def\algbackskip{\hskip-\ALG@thistlm}  
\makeatother

\author{Changxi You\thanks{C. You is 
		a Senior Researcher at Tencent Technology (Beijing) Company Limited, 100089, China. 	 	
		Email: {\small changxiyou@tencent.com}}%
}

\fancypagestyle{plain}{\fancyhead{}\fancyfoot[R]{\textbf{\thepage}}}
\fancypagestyle{empty}{\fancyhead{}\fancyfoot[R]{\textbf{\thepage}}}

\begin{document}

\maketitle

\pagestyle{empty}

\begin{abstract}
Collision avoidance is one of the most challenging tasks people need to consider for developing the self-driving technology.
In this paper we propose a new spatiotemporal motion planning algorithm that efficiently solves a constrained nonlinear optimal control problem using the iterative linear quadratic regulator (iLQR), which takes into account the uncertain driving behaviors of the traffic vehicles and minimizes the collision risks between the self-driving vehicle (referred to as the ``ego'' vehicle) and the traffic vehicles such that the ego vehicle is able to maintain sufficiently large distances to all the surrounding vehicles for achieving the desired collision avoidance maneuver in traffic.
To this end, we introduce the concept of the ``collision polygon'' for computing the minimum distances between the ego vehicle and the traffic vehicles, and provide two different solutions for designing the constraints of the motion planning problem by properly modeling the behaviors of the traffic vehicles in order to evaluate the collision risk. 
Finally, the iLQR motion planning algorithm is validated in multiple real-time tasks for collision avoidance using both a simulator and a level-3 autonomous driving test platform. \par


\end{abstract}

\section{Introduction}

Self-driving technique promises to partially or completely eliminate human driver behavioral error by reasonably rectifying the driver's inappropriate command during driving, which is expected to significantly enhance the vehicle's driving safety 
and represents a major trend in the future intelligent transportation systems.

Planning is one of the most important parts for developing the self-driving vehicles, which is mainly responsible for three tasks, including mission planning, where a routing problem is solved, decision making, where the vehicle takes an appropriate action according to certain strategy, and motion planning, where the vehicle plans its future trajectory as a function of space or time~\cite{Brechtel2011,Kuwata2009,Karaman2011}. In this paper we only concentrate on motion planning.
There are two different kinds of motion planning solutions in literature: 1) Path planning in the spatial domain with an additional speed profile generation in the temporal domain \cite{velenis2005optimal,you2019tcst}, where planning in space and planning in time are done sequentially in two separate steps. 2) A single solution where the trajectory is directly generated in the spatiotemporal domain, which is referred to as motion planning in this paper. 

Over the past decade numerous path planning algorithms have been developed using different methodologies, which can be categorized into three groups, namely, sampling \cite{Kuwata2009,Jaillet2010}, graph-search ~\cite{Garcia2013,Cimurs2017}, and geometry-based path planning~\cite{Choi2008,Choi2010,Shim2012}. 
In \cite{Kuwata2009} the authors proposed a real-time path planning algorithm based on Rapidly-exploring Random Trees (RRTs), which can efficiently explore the space to handle obstacle avoidance problems.  
Cimurs et al. used  Dijkstra's algorithm to find the shortest viable path by connecting the Vonoroi vertices, such that the path keeps a safe distance to all the obstacles in the environment \cite{Cimurs2017}. In \cite{wang2017real} a dynamic path planning algorithm based on D* is developed in order to avoid obstacles in complex environment involving multiple corners. All the above  mentioned path planning algorithms require an additional design of the path smoother to obtain the better continuity and smoothness. The geometry-based path planning methods may provide better smoothness by using a combination of arcs \cite{Chee1994,Fraichard2004}, clothoids\cite{Fraichard2004}, $\rm B\acute{e}zier$ \cite{chen2013lane,Korzeniowski2016,you2019autonomous} and polynomials\cite{Chee1994}. 

Motion planning in the spatiotemporal domain is a single solution that can simultaneously explore/exploit the spatial domain and the temporal domain for trajectory generation. Hence, it is expected to be able to satisfy the requirements for the most challenging autonomous driving tasks such as emergent collision avoidance and off-road high speed racing. Typical single-solution motion planning algorithms in the literature mainly include two categories: 1) sampling-based graph search \cite{kushleyev2009time,pivtoraiko2009diff,ziegler2009space,mcnaughton2011motion} and 2) trajectory optimization by solving optimal control problems\cite{chandra2017safe,obayashi2018real,chen2019auto,zhao2018,pan2020safe}.

Pivtoraiko et al. defined the so called ``state lattice'' using a searching graph by discretizing the state space for a moving robot, where the vertices represent the kinematic state of the robot and the edges represent the feasible motions for the transitions between the neighboring states \cite{pivtoraiko2009diff}. They successfully implemented both the A* \cite{hart1968formal} and the D* \cite{koenig2002d} search algorithms on a robot for off-road obstacles avoidance. The authors in \cite{mcnaughton2011motion} extended the state-lattice idea for autonomous driving in the structured environments, which was implemented in real-time on an autonomous passenger vehicle using a GPU. The sampling-based approaches are considered to be computationally expensive for evaluating the costs of a large number of the sampling trajectories \cite{gonzalez2015review}. 
Another direction for developing the spatiotemporal motion planning is related to the optimal control theory. Chandru et al. solved a model predictive control (MPC) problem for generating the safe lane change maneuvers in dense traffic \cite{chandra2017safe}. To better satisfy the constraints of the nonlinear vehicle dynamics for motion planning, Obayashi et al. formulated a nonlinear MPC (NMPC) problem for achieving more accurate motion planning \cite{obayashi2018real}. Nevertheless, the constrained NMPC is still a very hard problem to solve in general, which depends strongly on the system dynamics and the nonlinear constraints. Alternatively, an increasing number of the methods based on dynamic programming (DP) have been developed in the literature for solving the nonlinear optimization problems, such as the differential dynamic programming (DDP) \cite{mayne1973diff} and the iterative linear quadratic regulator (iLQR) \cite{chen2019auto,pan2020safe}, which may promise to provide better computation efficiency than NMPC. The authors in \cite{chen2019auto} proposed a constrained iLQR algorithm that efficiently solves the optimal control problem for motion planning under nonlinear constraints, which, indeed, shows some promising results on real-time collision avoidance.

\subsection*{Contributions}

This paper extends the iLQR motion planning algorithm from \cite{chen2019auto}, focusing on improving the ability of the autonomous vehicle in the dynamically changing environments for achieving the safe driving in different challenging tasks (i.e., emergent collision avoidance). To this end, we wish to accurately determine any possible overlap between the candidate future trajectory of the ego vehicle (EV) and the predicted trajectories of the surrounding traffic vehicles (TV), such that we can further refine the planned trajectory of the EV to ensure the driving safety over the next couple of seconds.

The main contribution of this paper is summarized as follows:
1) We propose to use a ``collision polygon'' to evaluate the collision risk in the design of the iLQR motion planning algorithm, where the collision polygon is defined using the Minkowski sum of the bounding polygons of the self-driving vehicle and the obstacles at each predicted time step. This collision polygon can be conveniently computed and it provides a more accurate result for detecting the possible overlap between the future trajectories of the EV and the TVs than using the approximating ellipses or circles in \cite{chen2017,pan2020safe}.
2) Next, we propose to use two methods for designing the constraints of the motion planning problem by considering the uncertain behaviors of the surrounding TVs, which, namely, include the minimum distance regulation (MDR) where we assume the TVs' behaviors can be exactly predicted, and the minimum risk regulation (MRR) where we assume the behaviors of the surrounding TVs are suffering from certain uncertainties.
To ease the computation work, a cone zone from the collision polygon is selected for efficiently computing the minimum distance between EV and the surrounding TVs. It is worth mentioning that in the MRR formulation we assume the predicted trajectory of each TV to be a Gaussian process, and design a sampling-based algorithm to evaluate the Gaussian integral over the cone zone (collision probability) and its first order and second order partial derivatives with respect to certain parameters (gradient and Hessian). 
As far as the authors know, we are the first to provide the detailed iLQR formulation that is able to handle the uncertainty of the predicted trajectories of the TVs by solving the Gaussian integral.
3) Instead of computing the collision probability by solving the Gaussian integrals subject to the arbitrary linear constraints, we further refine the MRR method by computing the expected cost using a barrier function of the distance between the EV and each TV having the uncertain predicted trajectories. This modification makes the iLQR motion planning algorithm possible for real-time implementation.
4) The approach in this paper is validated using both simulations and real-world field tests. Specifically, we implemented the iLQR algorithm in more than a hundred of manually designed test cases for emergent collision avoidance, which provides solid results to show both the efficiency and the effectiveness of the iLQR algorithm for the real-time collision avoidance of the self-driving vehicle in the dynamically-changing traffic environments.

The rest of the paper is organized as follows:  Section~\ref{sec:ocp} formulates the optimal control problem we want to solve for motion planning.
Section~\ref{sec:iLQR} briefly introduces the background theory of the iLQR approach. Section~\ref{sec:MotionPlanning} provides the detailed design of the cost functions and the constraints for motion planning.
Section~\ref{sec:result} validates the iLQR motion planning algorithm in a number of testing cases using both the simulation results and the real-world experiments. Finally, Section~\ref{sec:Conclusion} summarizes the results of this study.

\section{Problem Description}  \label{sec:ocp}
Given the initial state $x_{\rm start}$ of the EV, motion planning is the problem one solves for the desired future trajectory of the EV starting from $x_{\rm start}$ to a finite time horizon by maximizing (minimizing) certain reward (cost) with respect to certain constraints from the physical world.  By designing the cost functions and  the constraints, one is able to achieve the optimal driving maneuvers in different tasks by solving motion planning, such as lane-switching, on/off ramp merging and collision avoidance. Mathematically, this problem can be described as follows,
\begin{subequations} \label{eqn:ocp}
	\begin{align}
	\bold{x}^*, \bold{u}^* = \arg\, \min \limits_{\bold{x},\bold{u}} \, \Bigl\{&J(\bold{x},\bold{u}) = \ell^{\rm f}(x_{\rm T})+\sum_{k=0}^{T-1}\ell(x_k, u_k)\Bigr\}, \label{eqn:cost_function} \\
	{\rm s.t.} ~~~~~~~~~~~~ x_{k+1} &= f(x_k, u_k),  \label{eqn:ocp2} \\
	x_0 &= x_{\rm start}, \label{eqn:ocp3} \\
	{\rm and} ~~~~~~~~~~~~~~ g(&x_k, u_k)<0,  \label{eqn:ocp4} \\
	g^{\rm f}(x_{\rm T})<0,& ~~~~k=0,1,\cdots,T-1.  \label{eqn:ocp5}
	\end{align}
\end{subequations}
where $\bold{x}=\big\{x_k\big\}_{k=0,\cdots,T}$ and $\bold{u}=\big\{u_k\big\}_{k=0,\cdots,T}$ denote the state time series and the control time series, respectively; $\ell$ and $\ell^{\rm f}$ represent the process cost and the terminal cost, respectively; $g$ and $g^{\rm f}$ represent the process constraint and the terminal constraint, respectively. $k$ denotes the time step and $T$ denotes the total number of the steps for motion planning. The function $f(\cdot)$ in (\ref{eqn:ocp2}) represents the dynamics equation of the vehicle. 

\begin{figure}[H]
	\centering
	\includegraphics[width=7cm,height=2.5cm]{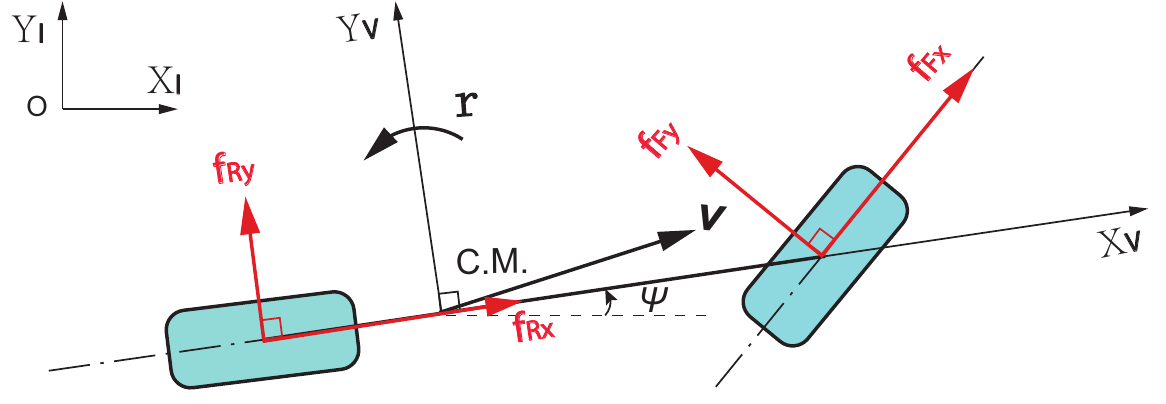}
	\caption{Single-track vehicle model.}
	\label{fig:vehicle}
\end{figure}

Fig.~\ref{fig:vehicle} shows the single-track vehicle dynamics model. $X_{\rm I}-O-Y_{\rm I}$ and $X_{\rm V}-\mathrm{CM}-Y_{\rm V}$ denote the inertial frame and the body-fixed frame, respectively.
Furthermore, $v$ denotes the velocity at the vehicle's center of mass (CM), and $\psi$ and $r$ denote the yaw angle and the yaw rate of the vehicle, respectively.
The symbols $f_{ij}$ ($i=F, R$ and $j=x, y$) denote the longitudinal and lateral friction forces at the front and rear wheels. 

The computations for the tire forces $f_{ij}$ ($i=F, R$ and $j=x, y$) are important for emergent collision avoidance, especially for the cases where the road friction conditions are poor. Nevertheless, sometimes one still have to sacrifice the vehicle modeling accuracy by ignoring the tire dynamics in order to simplify the mathematical problem to solve for the better real-time implementation speed. 
The feasibility of the solutions by solving such simplified motion planning problem can be guaranteed by properly designing the constraints considering the tire dynamics and the road friction condition.
Hence, we model the motion of the EV using only the kinematic equations and define the system state by $x=[p^{\rm x}, p^{\rm y}, v, \psi]^\mathrm{T}$, and define the control by $u=[a, r]^\mathrm{T}$, where $(p^{\rm x}, p^{\rm y})$ represent the coordinates of the CM of the EV in certain ground-fixed frame (i.e., $X_{\rm I}-O-Y_{\rm I}$), and $a$ denotes the longitudinal acceleration. The dynamics equation $f(\cdot)$ in (\ref{eqn:ocp2}) can be given by (\ref{eqn:eom1})-(\ref{eqn:eom4}),
\begin{subequations} \label{eqn:eom}
	\begin{align}
	p^{\rm x}_{k+1} &= p^{\rm x}_k + v_k\cos\psi\,{\rm dt}, \label{eqn:eom1}\\
	p^{\rm y}_{k+1} &= p^{\rm y}_k + v_k\sin\psi\,{\rm dt}, \\
	v_{k+1} &= v_k+a_k\,{\rm dt}, \\
	\psi_{k+1} &= \psi_k+r_k\,{\rm dt}, \label{eqn:eom4}
	\end{align}
\end{subequations}
where ${\rm dt}$ is the time interval used to discretize the system.

The most popular techniques for solving the nonlinear optimal control problem in (\ref{eqn:ocp}) may include the sequential quadratic programming (SQP) \cite{gill2005} and the interior-point optimizer (IPOPT) \cite{biegler2009}. Although both the SQP and the IPOPT have shown the powerful capability for solving the nonlinear constrained optimization problems, they are initially designed for the more general-purpose optimizations where the gradients are computed using finite difference, thus making them difficult to use for the real-time motion planning. This paper adopts the iLQR to solve the motion planning problem in (\ref{eqn:ocp}), since the iLQR is expected to be faster than both the IPOPT and the SQP \cite{chen2019auto}.

\section{Constrained Iterative LQR}  \label{sec:iLQR}
ILQR is a shooting-based method that optimizes the trajectory of the vehicle by tuning the control variables instead of directly tuning the trajectory itself. This approach is initially designed to solve the unconstrained optimization problems based on dynamic programming \cite{chen2017,todorov2005}. In this section we introduce the basic theory of the iLQR and the main idea of using the iLQR to solve the constrained motion planning problem.

\subsection{Basic Theory} \label{sec:Theory}
Let us ignore the constraints in (\ref{eqn:ocp4})-(\ref{eqn:ocp5}) for the moment and assume that we want to solve only the unconstrained optimization problem in (\ref{eqn:cost_function})-(\ref{eqn:ocp3}). The iLQR then computes the optimal control strategy for the $k\,$th time step using dynamic programming following the Bellman equation,
\begin{align} \label{eqn:value}
 V(x_k) = \min \limits_{u_k} \Big[ \ell(x_k, u_k)+V(x_{k+1}) \Big],
\end{align}
where $V(\cdot)$ is the value function that represents the minimum cost-to-go starting at the current state $x_k$. We then define the state-control value function $Q$ as follows,
\begin{align}
Q(x_k,u_k) &= \ell(x_k, u_k)+V(x_{k+1}), \nonumber \\
		   &= \ell(x_k, u_k)+V(f(x_k, u_k)).
\end{align}
In order to derive for the optimal control strategy, we need to perturb $Q(\cdot)$ around $(x_k,u_k)$ as follows,
\begin{align}
P(\delta x_k,\delta u_k) = Q(x_k+\delta x_k,u_k+\delta u_k)-Q(x_k,u_k), 
\end{align}
where $P(\cdot)$ denotes the perturbation and $(\delta x_k,\delta u_k)$ denote the small variations in the state and the control, respectively. 
Next, we quadratize the perturbation function $P(\cdot)$ around $(0,0)$, which can be approximated by its second-order Taylor expansion,
\begin{align}  \label{eqn:P_quadratic}
P(\delta x_k,\delta u_k)\approx\frac{1}{2}
\left[ 
{\begin{array}{*{20}c}
	1\\	\delta x_k\\ \delta u_k\\
	\end{array}}
\right]^\mathrm{T}
\left[ 
{\begin{array}{*{20}c}
	0  & P_x^\mathrm{T} & P_u^\mathrm{T}\\
	P_x  & P_{xx} & P_{xu}\\
	P_u  & P_{ux} & P_{uu}\\	
\end{array}}
\right]
\left[ 
{\begin{array}{*{20}c}
	1\\	\delta x_k\\ \delta u_k\\
	\end{array}}
\right],
\end{align}
where the subscripts of $P$ denote the first and second order partial derivatives of $P(\cdot)$, which are given by
\begin{subequations} \label{eqn:partials}
	\begin{align}
	P_x &= \frac{\partial \ell}{\partial x}\Bigg|_{x_k,u_k}
	+\frac{\partial f}{\partial x}\Bigg|_{x_k,u_k}^\mathrm{T}
	\cdot \frac{\partial V}{\partial x}\Bigg|_{x_{k+1},u_{k+1}}
	\triangleq \ell_x + f_x^\mathrm{T}V'_x, \label{eqn:partials1}\\
	P_u &= \frac{\partial \ell}{\partial u}\Bigg|_{x_k,u_k}
	+\frac{\partial f}{\partial u}\Bigg|_{x_k,u_k}^\mathrm{T}
	\cdot \frac{\partial V}{\partial x}\Bigg|_{x_{k+1},u_{k+1}}
	\triangleq \ell_u + f_u^\mathrm{T}V'_x, \label{eqn:partials2}\\
	P_{xx} &= \frac{\partial^2 \ell}{\partial x^2}\Bigg|_{x_k,u_k}
	+\frac{\partial f}{\partial x}\Bigg|_{x_k,u_k}^\mathrm{T}
	\cdot \frac{\partial^2 V}{\partial x^2}\Bigg|_{x_{k+1},u_{k+1}}
	\cdot \frac{\partial f}{\partial x}\Bigg|_{x_k,u_k} \nonumber \\
	+& \frac{\partial V}{\partial x}\Bigg|_{x_{k+1},u_{k+1}}
	\cdot \frac{\partial^2 f}{\partial x^2}\Bigg|_{x_k,u_k}
	\triangleq \ell_{xx} + f_x^\mathrm{T}V'_{xx}f_x + V'_xf_{xx}, \label{eqn:partials3}\\
	P_{uu} &= \frac{\partial^2 \ell}{\partial u^2}\Bigg|_{x_k,u_k}
	+\frac{\partial f}{\partial u}\Bigg|_{x_k,u_k}^\mathrm{T}
	\cdot \frac{\partial^2 V}{\partial x^2}\Bigg|_{x_{k+1},u_{k+1}}
	\cdot \frac{\partial f}{\partial u}\Bigg|_{x_k,u_k} \nonumber \\
	+& \frac{\partial V}{\partial x}\Bigg|_{x_{k+1},u_{k+1}}
	\cdot \frac{\partial^2 f}{\partial u^2}\Bigg|_{x_k,u_k}
	\triangleq \ell_{uu} + f_u^\mathrm{T}V'_{xx}f_u + V'_xf_{uu}, \label{eqn:partials4}\\	
	P_{xu} &= P_{ux}^\mathrm{T} =
	\frac{\partial^2 \ell}{\partial x \partial u}\Bigg|_{x_k,u_k}
	+\frac{\partial f}{\partial x}\Bigg|_{x_k,u_k}^\mathrm{T}
	\cdot \frac{\partial^2 V}{\partial x^2}\Bigg|_{x_{k+1},u_{k+1}}
	\cdot \frac{\partial f}{\partial u}\Bigg|_{x_k,u_k} \nonumber \\
	+& \frac{\partial V}{\partial x}\Bigg|_{x_{k+1},u_{k+1}}
	\cdot \frac{\partial^2 f}{\partial x \partial u}\Bigg|_{x_k,u_k}
	\triangleq \ell_{xu} + f_x^\mathrm{T}V'_{xx}f_u + V'_xf_{xu}. \label{eqn:partials5}
	\end{align}
\end{subequations}
In order to ease the computation, a common simplification for the iLQR is to eliminate the second-order terms $f_{xx}$, $f_{xu}$ and $f_{uu}$ in (\ref{eqn:partials3})-(\ref{eqn:partials5}).  Hence, the optimal control variation can be determined by minimizing $P(\delta x_k,\delta u_k)$,
\begin{align}
\delta u_k^* = \arg\, \min \limits_{\delta u_k} P(\delta x_k,\delta u_k).
\end{align}
We then let $\partial P(\delta x_k,\delta u_k)/\partial (\delta u_k) = P_u+P_{ux}\delta x + P_{uu}\delta u= 0$ and obtain the following equation,
\begin{align} \label{eqn:u}
\delta u_k^* = H + G\cdot\delta x_k, 
\end{align}
where $H=-P_{uu}^{-1}P_u$ and $G=-P_{uu}^{-1}P_{ux}$.
The first and second order partial derivatives of the value function $V(\cdot)$ with respect to $x$ are given by
\begin{subequations} \label{eqn:value_x}
\begin{align}
V_x &= P_x-P_uP_{uu}^{-1}P_{ux}, \\
V_{xx} &= P_{xx}-P_{xu}P_{uu}^{-1}P_{ux}.
\end{align}
\end{subequations}

\begin{figure}[!htbp]
	\centering
	\includegraphics[width=0.5\textwidth,height=0.17\textwidth]{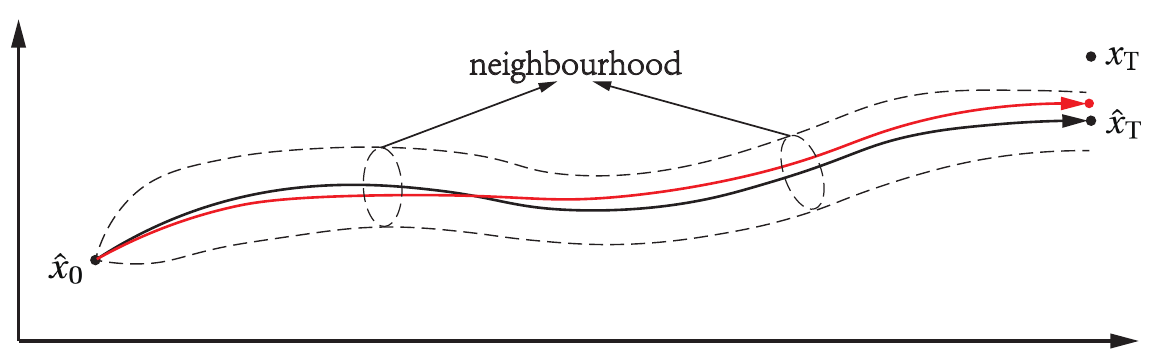}
	\caption{iLQR trajectory optimization in a single iteration.}
	\label{fig:1_ilqr}
\end{figure}

The results in (\ref{eqn:partials})-(\ref{eqn:value_x}) indicate that if certain reference state sequence $\hat{\bold{x}}=\big\{\hat{x}_k\big\}_{k=0,\cdots,T}$ and the corresponding  control sequence  $\hat{\bold{u}}=\big\{\hat{u}_k\big\}_{k=0,\cdots,T}$ are provided, one is able to quadratize the system $f(\cdot)$ and the cost $\ell(\cdot)$ about the reference $(\hat{x}_k,\hat{u}_k)$ from $k=T$ to $k=0$ and compute the optimal control variations $\delta u_k^*$ for each time step $k$ using the dynamic programming method. Hence, the essential part of one single iLQR iteration contains a pair of the forward propagation and the backward propagation processes, where the forward propagation evaluates the control sequence $\hat{\bold{u}}$ over the planning time horizon and updates the value function $V(\cdot)$ along the planned trajectory $\hat{\bold{x}}$, and the backward propagation computes the control update gains $H$ and $G$ at each time step and slightly adjusts the control sequence $\hat{\bold{u}}$ by adding the optimal control variations $\delta u^*$ in order to minimize the cost-to-go $V(\cdot)$. Fig.~\ref{fig:1_ilqr} graphically shows this idea, where the black curve $\big\{\hat{x}_k\big\}_{k=0\cdots,T}$ represents the reference trajectory before optimization, and the red curve represents the newly generated trajectory after one single iLQR iteration. The symbol $x_{\rm T}$ in the figure denotes the target state. Such processes can be repeated for a number of iterations until certain desired trajectory is found.

\subsection{Barrier Function} \label{sec:barrier}
We recall the fact that the constraints in (\ref{eqn:ocp4})-(\ref{eqn:ocp5})  have been ignored at the beginning of Section~\ref{sec:Theory}.
The most popular idea to solve the constrained optimization problem in (\ref{eqn:cost_function})-(\ref{eqn:ocp5}) using the iLQR is to relax the hard constraints by transforming them into a couple of the cost terms using different kinds of barrier functions, such as the exponential barrier function \cite{chen2017,pan2020safe} and the log barrier function \cite{chen2019auto}. 
This method is referred to as the constrained iLQR in the literature \cite{todorov2005,chen2017,chen2019auto,pan2020safe}.

In this paper we select to use the exponential barrier function instead of the log barrier function, since the log barrier function is not defined on the negative domain, which may cause the infeasibility problem during the implementation of the iLQR. 
Given the constraint function $g(x_k, u_k)<0$ in (\ref{eqn:ocp4}), for instance, the new cost term transformed using the exponential barrier function can be defined as follows,
\begin{align} \label{eqn:exp}
b(x_k, u_k) = q_1 \exp \big(q_2g(x_k, u_k)\big),
\end{align}
where $q_1>0$, $q_2>0$ are the coefficients to be designed. Furthermore, equation (\ref{eqn:exp}) can be quadratized as follows,
\begin{align} \label{eqn:b_quadratic}
&b(x_k+\delta x_k, u_k+\delta u_k) \approx \nonumber\\
&b(x_k, u_k) +
\frac{1}{2}
\left[ 
{\begin{array}{*{20}c}
	1\\	\delta x_k\\ \delta u_k\\
	\end{array}}
\right]^\mathrm{T}
\left[ 
{\begin{array}{*{20}c}
	0  & b_x^\mathrm{T} & b_u^\mathrm{T}\\
	b_x  & b_{xx} & b_{xu}\\
	b_u  & b_{ux} & b_{uu}\\	
	\end{array}}
\right]
\left[ 
{\begin{array}{*{20}c}
	1\\	\delta x_k\\ \delta u_k\\
	\end{array}}
\right],
\end{align}
where 
\begin{subequations} \label{eqn:b_partials}
	\begin{align}
	b_x &= q_1q_2\exp\big(q_2g(x_k, u_k)\big)g_x, \label{eqn:b_partials1}\\
	b_u &= q_1q_2\exp\big(q_2g(x_k, u_k)\big)g_u, \label{eqn:b_partials2}\\
	b_{xx} &= q_1q_2\exp\big(q_2g(x_k, u_k)\big)\big(q_2g_xg^{\mathrm{T}}_x+g_{xx}\big), \label{eqn:b_partials3}\\
	b_{uu} &= q_1q_2\exp\big(q_2g(x_k, u_k)\big)\big(q_2g_ug^{\mathrm{T}}_u+g_{uu}\big), \label{eqn:b_partials4}\\	
	b_{xu} &= q_1q_2\exp\big(q_2g(x_k, u_k)\big)\big(q_2g_xg^{\mathrm{T}}_u+g_{xu}\big), \label{eqn:b_partials5}
	\end{align}
\end{subequations}
and where the subscripts of $g(\cdot)$ represent the first order and the second order partial derivatives of the constraint function. The equations in (\ref{eqn:b_partials1})-(\ref{eqn:b_partials5}) provide the results for the gradients and the Hessian matrices of the transformed cost function in (\ref{eqn:exp}). More results and discussions on the constrained iLQR can be found in \cite{chen2017}.

%
%

\section{Motion Planning}   \label{sec:MotionPlanning}

In this section we design the cost functions and the constraints for the motion planning problem in (\ref{eqn:cost_function})-(\ref{eqn:eom4}). Specifically, we provide two different solutions for collision avoidance: 1) The state estimations and the predicted trajectories for the TVs are provided in high confidence. In this case one is able to trust the quality of the predicted trajectories and further ensure the driving safety by maintaining the distances between the EV and the TVs to be larger than certain safe distance since the distances between the EV and the TVs can be exactly computed. 2) The state estimations and the predicted trajectories for the TVs are provided with considerable uncertainties. Instead of directly maintaining the distances between the EV and the TVs, in this case we penalize the collision probabilities between the EV and the TVs for the enhanced driving safety. 

In order to simplify the mathematical expressions for the remaining part of this section, we use $\boldsymbol{I}_{n}$ and $\boldsymbol{O}_{a\times b}$ to represent the identity matrices and the zero matrices, respectively, where $n$, $a$ and $b$ denote the dimensions for the identity matrices and the zero matrices, respectively.
We first introduce the design of the cost function.

\subsection{Cost Function}
We recall the fact that the optimal control variations $\delta u^*$ in (\ref{eqn:u}) are computed using the gradients and the Hessian matrices of the cost function $\ell$. Hence, it is convenient to design $\ell$ in the form of the sum of a series of quadratic expressions to ease the computation.

\subsubsection{Control Cost}
The control cost is designed to reduce the control energy that is required for achieving the target trajectory. Moreover, one can also minimize the deviation of the control signal from its desired value for providing the better smoothness of the trajectory in both the longitudinal and the lateral directions. We design the control cost as follows,
\begin{align} \label{eqn:control}
\ell_u = \frac{1}{2}u_k^\mathrm{T}
\left[ 
{\begin{array}{*{20}c}
	w_{\rm a} & 0\\ 0 & w_{\rm r}\\
	\end{array}}
\right]
u_k,
\end{align}
where $w_{\rm a}$ and $w_{\rm r}$ are the weights to be specified.
\subsubsection{Adjusting Cost}
Next, we introduce the adjusting cost. Given the reference trajectory  $\hat{\bold{x}}\triangleq\{\hat{x}_k\}_{k=0\cdots,T}$ (see Fig.~\ref{fig:1_ilqr}) in each iLQR iteration, where $\hat{\bold{x}}$ can be obtained from the previous iteration, one may want to optimize the trajectory by gradually adjusting $\hat{\bold{x}}$ within its certain neighborhood such that the new trajectory cannot be obviously changed from the reference $\hat{\bold{x}}$ in one iteration. Such designed cost is referred to as the ``adjusting cost'' in this work, which is used to mitigate the dynamic effect of the modeling errors by the system linearizations about different references $\hat{\bold{x}}$ in different iterations. The adjusting cost is given by
\begin{align}
\ell_{\hat{\bold{x}}} = \frac{1}{2}(x_k-\hat{x}_k)^\mathrm{T}
\left[ 
{\begin{array}{*{20}c}
	w_{\rm p^x} & 0 & 0 & 0 \\ 0& w_{\rm p^y} & 0 & 0 \\
	0 & 0 & w_{\rm v} & 0 \\ 0 & 0 & 0 & w_{\rm \psi} \\
	\end{array}}
\right]
(x_k-\hat{x}_k),
\end{align}  
where  $w_{\rm p^x}$, $w_{\rm p^y}$, $w_{\rm v}$ and $w_{\rm \psi}$ are the weights to be specified.

\begin{figure}[!htbp]
	\centering
	\includegraphics[width=0.5\textwidth,height=0.165\textwidth]{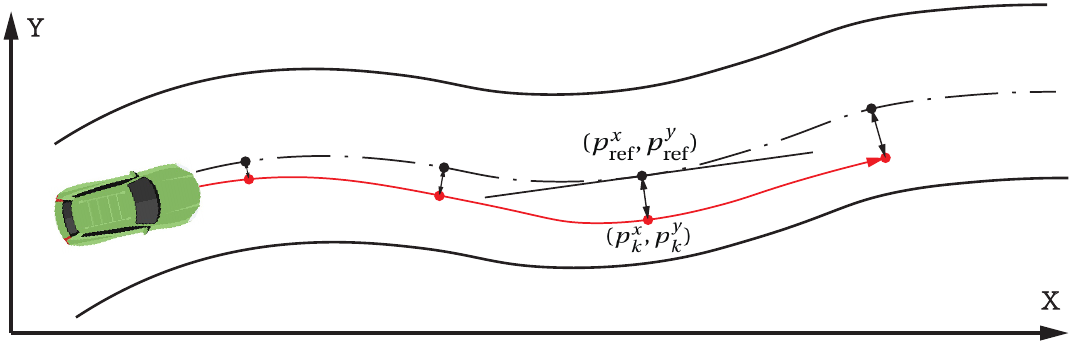}
	\caption{Tracking error for lane keeping.}
	\label{fig:2_tracking}
\end{figure}

\subsubsection{Tracking Cost }
Tracking cost is usually designed for achieving better lane-keeping. To this end, for each point ${(p^x_k,p^y_k)}$ we first find its reference point ${(p^x_{\rm ref},p^y_{\rm ref})}$, which is the closest point to ${(p^x_k,p^y_k)}$ on the reference line (i.e., road center-line in Fig.~\ref{fig:2_tracking}). Moreover, one can also specify the desired speed $v_{\rm ref}$ accordingly. Then, the tracking cost can be determined as follows,
\begin{align} \label{eqn:tracking_cost}
\ell_{\rm ref} = \frac{1}{2}\Big(Cx_k-x_{\rm ref}\Big)^\mathrm{T}
\left[ 
{\begin{array}{*{40}c}
	w_{\rm p^{\rm ref}} & 0 & 0  \\ 
	0& w_{\rm p^{\rm ref}} & 0  \\
	0 & 0 & w_{\rm v^{\rm ref}}  \\ 
	\end{array}}
\right]
\Big(Cx_k-x_{\rm ref}\Big),
\end{align}
where $w_{\rm p^{\rm ref}}$ and $w_{\rm v^{\rm ref}}$ are the weights to be specified, $x_{\rm ref}=\big[p^x_{\rm ref},p^y_{\rm ref}, v_{\rm ref}\big]^\mathrm{T}$, and the matrix
$C=\left[ {\begin{array}{*{10}c} \boldsymbol{I}_3 & |& \boldsymbol{O}_{3\times 1} \\ \end{array}} \right]$.
\subsubsection{Terminal Cost}
Similar with the design of the tracking cost in (\ref{eqn:tracking_cost}), we design the terminal cost to achieve the desired orientation $\psi^{\rm f}$ and the desired speed $v^{\rm f}$ at $k=T$, which can be given by
\begin{align} \label{eqn:terminal}
\ell^{\rm f} = \frac{1}{2}\Big(C^{\rm f}x_{\rm T}-x^{\rm f}\Big)^\mathrm{T}
\left[ 
{\begin{array}{*{40}c}
	w^{\rm f}_{\rm \psi} & 0   \\ 
	0& w^{\rm f}_{\rm v} \\
	\end{array}}
\right]
\Big(C^{\rm f}x_{\rm T}-x^{\rm f}\Big),
\end{align}
where $w^{\rm f}_{\rm \psi}$ and $w^{\rm f}_{\rm v}$ are the weights to be specified, $x^{\rm f}=\big[\psi^{\rm f}, v^{\rm f}\big]^\mathrm{T}$, and the matrix $C^{\rm f}=\left[ {\begin{array}{*{10}c} \boldsymbol{O}_{2\times 2} & |& \boldsymbol{I}_2 \\ \end{array}} \right]$.

\subsection{Constraint Function}
In this section we define the constraints for motion planning. The constraint functions are used to describe certain feasible space in which the vehicle is allowed to drive, which are designed mainly by considering the safety reasons and other restrictions from the physical world.

\subsubsection{Control Constraint}
The constraints on control are represented by the following expression,
\begin{align} \label{eqn:u_cons}
\left[ {\begin{array}{*{40}c}
	a_{\rm min} \\ 
	r_{\rm min} \\
	\end{array}}
\right] 
\prec u_k \prec
\left[ {\begin{array}{*{40}c}
	a_{\rm max} \\ 
	r_{\rm max} \\
	\end{array}}
\right] 
\end{align}
where $a_{\rm min}$ and $r_{\rm min}$ denote the minimum acceleration and the minimum yaw rate, respectively; $a_{\rm max}$ and $r_{\rm max}$ denote the maximum acceleration and the maximum yaw rate, respectively. The symbol ``$\prec$'' represents component-wise inequalities. The lower and upper bounds for $u_k$ need to be carefully designed using certain reasonable values to guarantee the feasibility of such trajectory for tracking control in the next step.
We denote the peak road friction coefficient as $\hat{\mu}$. Then, the following inequalities must hold,
\begin{subequations}
\begin{align}
-\hat{\mu} g<a_{\rm min} &< a_{\rm max} < \hat{\mu} g, \\
-\sqrt{\hat{\mu}^2g^2-\overline{a}^2}\big/C_{\rm v}x_k < r_{\rm min} &< r_{\rm max} < \sqrt{\hat{\mu}^2g^2-\overline{a}^2}\big/C_{\rm v}x_k,
\end{align}
\end{subequations} 
where $\overline{a}=\max\big\{\,|a_{\rm min}|,|a_{\rm max}|\,\big\}$, $C_{\rm v}=[0,0,1,0]$, and $g = 9.81\, [{\rm m/s^2}]$ is the gravitational acceleration constant.
\subsubsection{Boundary Constraint}
The tracking cost in ($\ref{eqn:tracking_cost}$) penalizes the deviation of the planned future trajectory from the lane center-line. Nevertheless, one still needs to consider the road boundary constraints for motion planning in order to avoid possible severe collisions with the static barriers along the boundaries. One convenient way to define these boundary constraints is to represent the boundary curves using certain order polynomial function in an appropriate frame (see Fig.~\ref{fig:2_tracking}). The boundary constraints can be described as follows,
\begin{subequations} \label{eqn:boundaries}
\begin{gather}
C_{\rm y}x_k - \Gamma_{\rm left}\big(C_{\rm x}x_k \big)<0, \\
C_{\rm y}x_k - \Gamma_{\rm right}\big(C_{\rm x}x_k \big)>0,
\end{gather}
\end{subequations} 
where $C_{\rm x}=[1,0,0,0]$, $C_{\rm y}=[0,1,0,0]$, and $\Gamma_{\rm left}(\cdot)$ and $\Gamma_{\rm right}(\cdot)$ represent the left and right boundary curves, respectively, which are given by
\begin{gather} 
\Gamma_{\rm left}(x) = \sum_{k=0}^{n_1} a_kx^k, \qquad \Gamma_{\rm right}(x) = \sum_{k=0}^{n_2} b_kx^k,
\end{gather}
where $n_1$ and $n_2$ are positive integers, and $a_k$ and $b_k$ are the coefficients.

\begin{figure}[!htbp]
	\centering
	\includegraphics[width=0.4\textwidth,height=0.3\textwidth]{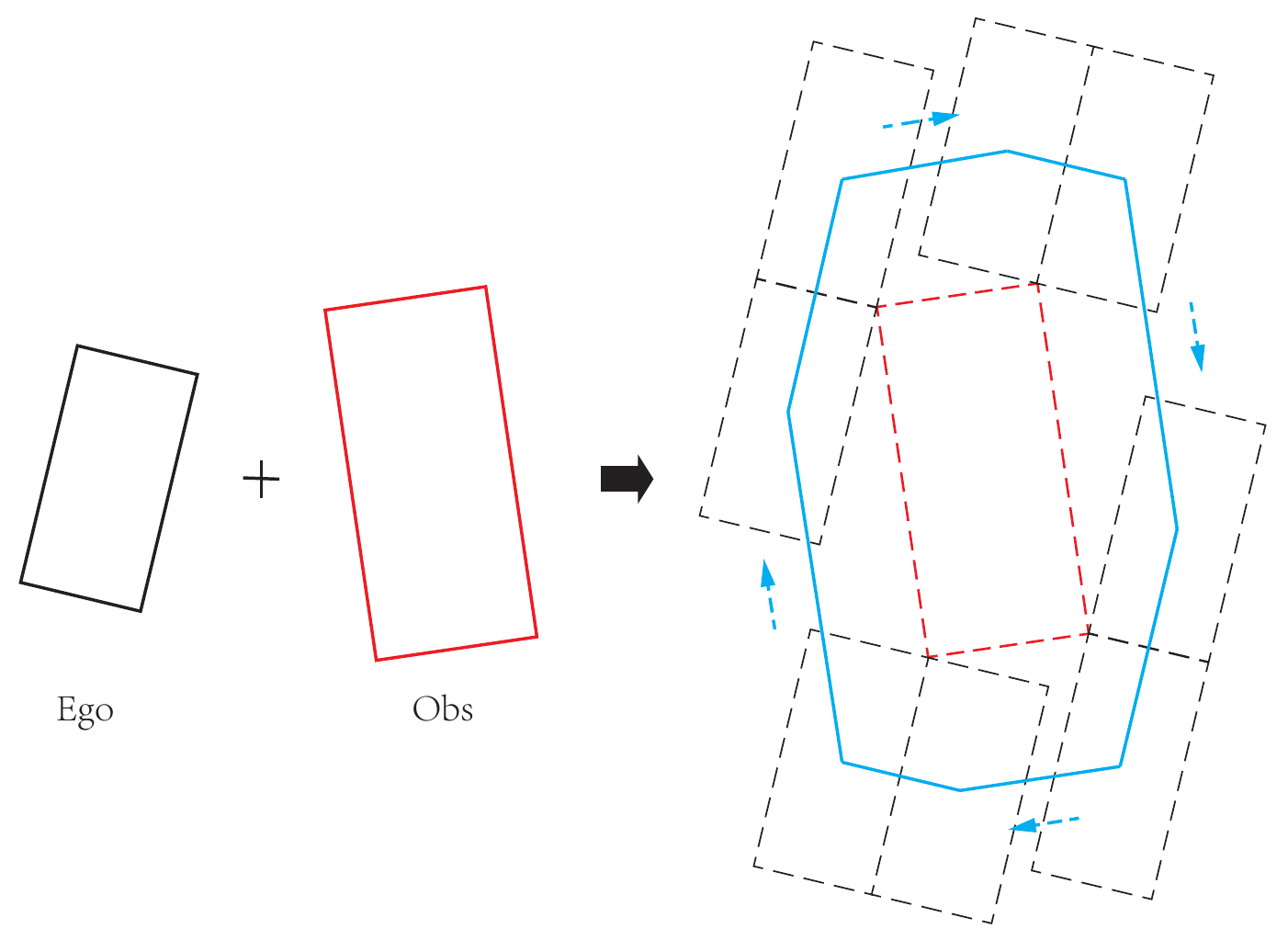}
	\caption{Minkowski sum of the bounding polygons of EV and a TV.}
	\label{fig:3_octagon}
\end{figure}

\subsubsection{Obstacle Constraint}
The design of the obstacle constraints is the most important task for motion planning using iLQR, which dynamically evaluate the potential collision risk between EV and the surrounding TVs and rectify therefore the EV's future trajectory planning by decreasing the collision risk. We assume that the trajectories of the TVs are predicted with different levels of uncertainties and provide two different solutions accordingly.

Let us assume there are $M$ surrounding TVs to be considered and define the TV set $\mathcal{S}\triangleq \big\{{\rm TV}_j\big|j=1,\cdots,M\big\}$. Then, we represent each ${\rm TV}_j$ in $\mathcal{S}$ using a bounding polygon in the two-dimensional space, where the bounding polygon is the convex hull of ${\rm TV}_j$ defined using a number of vertices $V_1,\cdots,V_L$,
\begin{align}
 \mathcal{B}_j\triangleq\Big\{\sum_{i=0}^{L}\rho_iV_i\Big|V_i\in\mathbb{R}^2,\rho_i\geq0, \sum_{i=0}^{L}\rho_i=1\Big\}, ~ j=1,\cdots,M.
\end{align} 
For the sake of simplicity we let $L=4$ and use the bounding rectangles to approximately represent the TVs.
Then, the collision polygon for ${\rm TV}_j$ can be defined using the convolution of the oriented bounding rectangle of the EV and the oriented bounding rectangle of ${\rm TV}_j$ as follows (see Fig.~\ref{fig:3_octagon}),
\begin{align}
\mathcal{C}_j\triangleq \big\{P'+P''-P\big|P'\in\mathcal{B},P''\in\mathcal{B}_{j}\big\}, ~ j=1,\cdots,M,
\end{align} 
where $P$ and $\mathcal{B}$ denote the centroid and the bounding rectangle of the EV, respectively. The boundary of $\mathcal{C}_j$ can be viewed as the trace of the EV's centriod when the EV moves around the border of $\mathcal{B}_{j}$. We further represent the predicted trajectory for ${\rm TV}_j$ as 
\begin{align}
\mathcal{Z}_j\triangleq\Big\{\zeta^k_j\in \mathbb{R}^2 \Big|k=0,\cdots,T\Big\},
\end{align}
where $\zeta^k_j$ denotes the position of the centroid for ${\rm TV}_j$ at the $k$\,th prediction time step.

Next, we design the constraints for obstacle avoidance based on the collision polygons $\mathcal{C}_j$ using two different approaches, namely, minimum distance regulation (MDR) and minimum risk regulation (MRR).

\begin{figure}[!htbp]
	\centering
	\includegraphics[width=0.3\textwidth,height=0.38\textwidth]{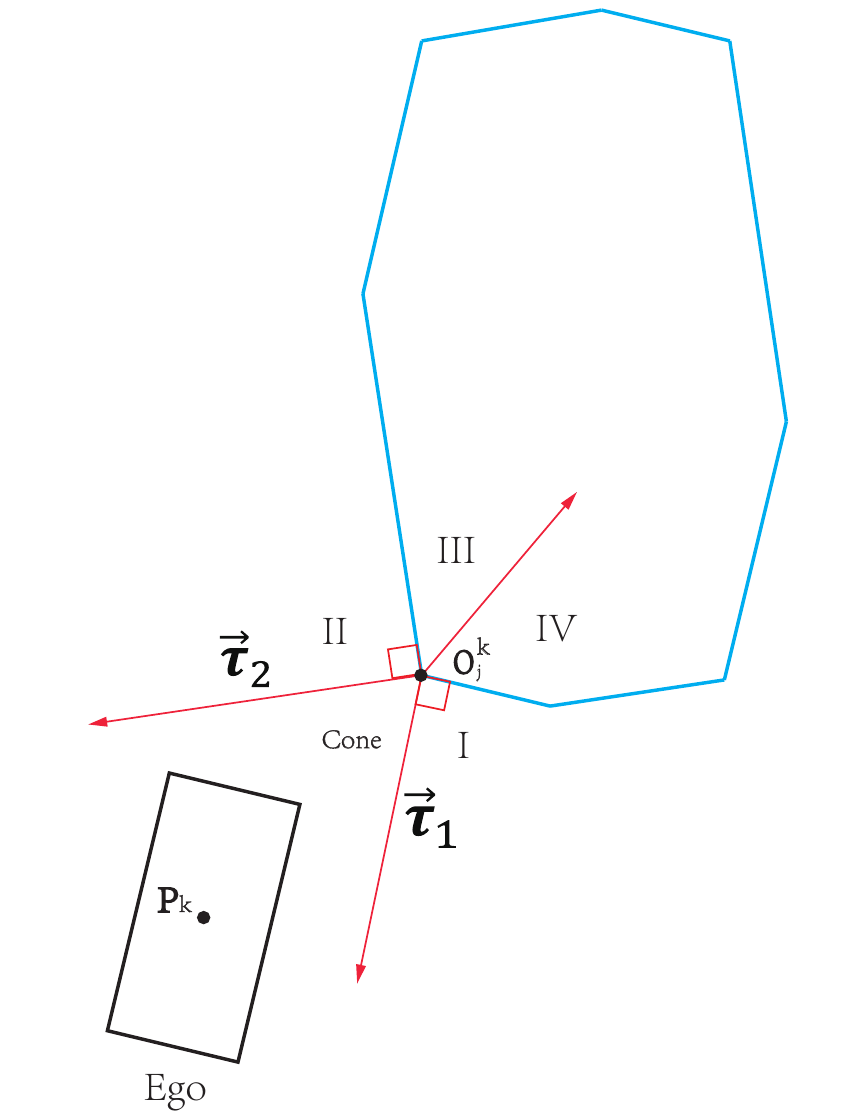}
	\caption{Distance between EV and ${\rm TV}_j$ at the $k$\,th prediction time step.}
	\label{fig:4_octagon_dis}
\end{figure}

a) \textit{MDR}. If the prediction trajectory for ${\rm TV}_j$ can be provided with sufficient accuracy, one just needs to determine the distance $d^k_j$ between the EV and ${\rm TV}_j$ for each predicted time step $k$ and maintain $d^k_j$ to be larger than certain reasonable value in order to ensure the driving safety. 

We denote the collision polygon of ${\rm TV}_j$ at the time step $k$ as $\mathcal{C}_j^k$. Then, we determine the closest vertex of the collision polygon $\mathcal{C}_j^k$ to the EV's centriod $P_k$, which is indicated by the point $O_j^k$ in Fig.~\ref{fig:4_octagon_dis}.
Next, we divide the two dimensional space around $O_j^k$ into five zones, including the zone I, II, III, IV and a cone zone, where the zone III and zone IV are divided by the bisector of the angle $\angle O_j^k$. The minimum distance between the EV and the ${\rm TV}_j$ at the time step $k$ can be computed as follows,
\begin{align} \label{eqn:dis}
d^k_j\big(x_k, O_j^k\big)=\left\{ {\begin{array}{*{20}l} 
\overrightarrow{\tau}_1\cdot \overrightarrow{O_j^kP_k},\qquad ~ P_k\in {\rm I}\cup{\rm  IV}, \\ 
\overrightarrow{\tau}_2\cdot \overrightarrow{O_j^kP_k},\qquad ~ P_k\in {\rm II}\cup{\rm  III}, \\ 
\|\overrightarrow{O_j^kP_k}\|,\qquad \quad ~ P_k\in {\rm Cone},\\ 
	\end{array}}\right. 
\end{align}
where $\overrightarrow{\tau}_1$ and $\overrightarrow{\tau}_2$ are the two normal vectors perpendicular to the two edges of $\angle O_j^k$, respectively; $P_k$ denotes the centroid of the EV. Furthermore, the obstacle constraints can be given by
\begin{align} \label{eqn:TV_cons1}
d^k_j\big(x_k, O_j^k\big) - \underline{d}>0, \qquad j = 1,\cdots, M,
\end{align}
where $\underline{d}>0$ is the minimum required distance between EV and the TVs.

\begin{figure}[!htbp]
	\centering
	\includegraphics[width=0.35\textwidth,height=0.275\textwidth]{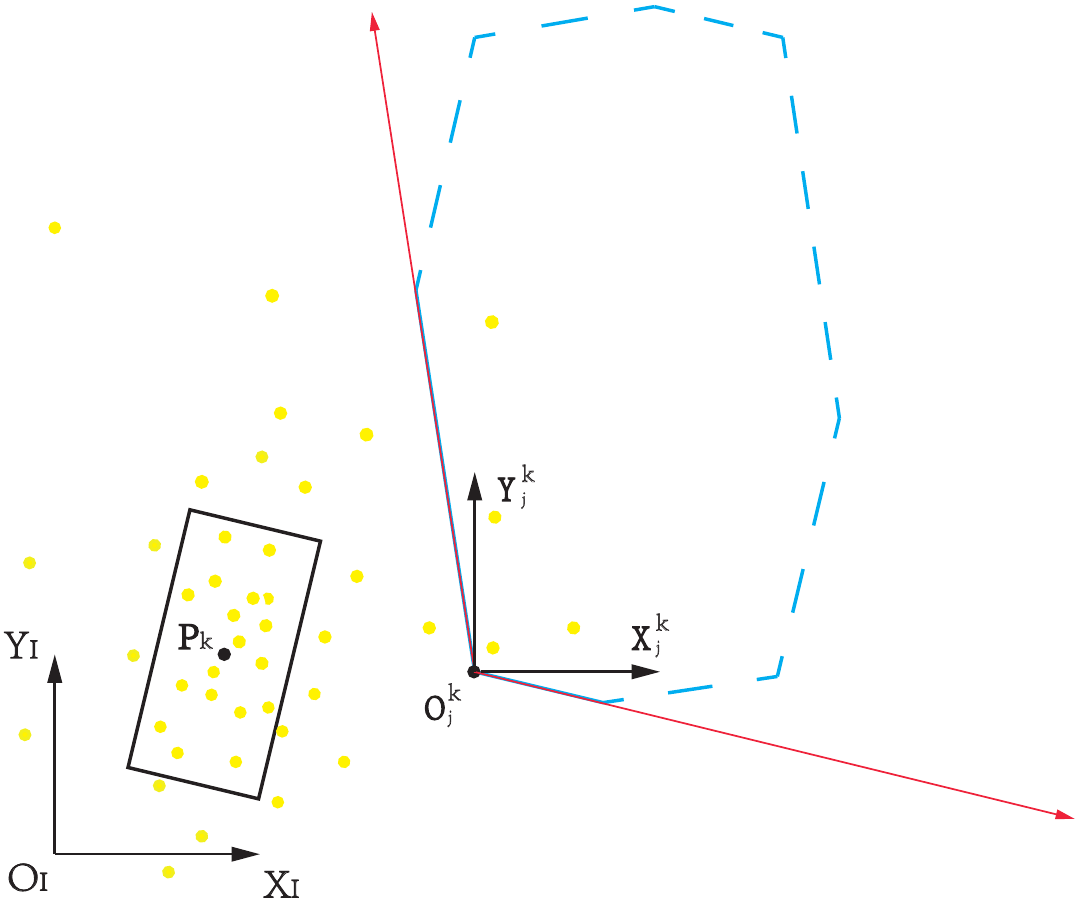}
	\caption{The collision probability of EV and ${\rm TV}_j$ at the $k$\,th prediction time step.}
	\label{fig:5_octagon_prob}
\end{figure}

b) \textit{MRR}. The constraints in (\ref{eqn:TV_cons1}) are only valid when the uncertainties associated with both the state estimations and the trajectory predictions for the TVs can be neglected, but in practice this condition cannot be always satisfied. In the cases where the TVs' states are poorly measured or their future trajectories are predicted with significant uncertainties, one still needs to take into account these uncertainties for safe motion planning. 

We assume the orientation of each TV is in parallel with the tangential direction of its predicted trajectory over the prediction horizon. Thus, the variance of the orientation of each TV is considered to be very low and hence negligible \cite{philipp2019}. Consequently, we only consider the covariance of the position vector for each TV in order to model the uncertainties for the TVs' behaviors. We further neglect the covariance of the EV's position vector based on the fact that the state of the EV can be measured more accurately than the surrounding TVs.

Instead of computing the distance from the centroid of the EV $P_k$ to the collision polygon $\mathcal{C}_j^k$ using (\ref{eqn:dis}), one needs to compute the probability of $P_k$ to be in $\mathcal{C}_j^k$ since the position of $O_j^k$ is no longer deterministic (see Fig.~\ref{fig:4_octagon_dis}). Furthermore, in order to simplify the derivations, we suggest to use the cone formed by $\angle O_j^k$ instead of using the polygon $\mathcal{C}_j^k$ itself to approximately calculate the collision probability as shown in Fig.~\ref{fig:5_octagon_prob}. To avoid the ambiguities in the notations, in the remaining part we use ``$\measuredangle$'' to denote a cone and use ``$\angle$'' to denote an angle, respectively.

Let us assume $O_j^k \sim \mathcal{N}(\mu_j^k,\Sigma_j^k)$ and define a new random vector $\tilde{P}_k\triangleq \overrightarrow{O_j^kP_k} =C_{\rm P}x_k-O_j^k$, where $C_{\rm P}=\left[ {\begin{array}{*{10}c} \boldsymbol{I}_2 & |& \boldsymbol{O}_{2\times 2} \\ \end{array}} \right]$.
It is not hard to figure out $\tilde{P}_k\sim \mathcal{N}(C_{\rm P}x_k-\mu_j^k,\Sigma_j^k)$. Then, we just need to compute the probability of the random point $\tilde{P}_k$ observed in the frame $X_j^k-O_j^k-Y_j^k$ (see Fig.~\ref{fig:5_octagon_prob}) to be in the cone $\measuredangle O_j^k$, which can be given as follows,
\begin{align} \label{eqn:gaussian}
&\mathbb{P}\Big(\tilde{P}_k\in \,\measuredangle O_j^k\Big)=\int\displaylimits_{\measuredangle O_j^k}\frac{1}{2\pi\sqrt{|\Sigma_j^k|}}\exp\Bigg(-\frac{1}{2}\Big(\xi-\mu_j^k+C_{\rm P}x_k\Big)^\mathrm{T}\Big(\Sigma_j^k\Big)^{-1} \nonumber \\
&~\,\Big(\xi-\mu_j^k+C_{\rm P}x_k\Big)\Bigg)\, {\rm d}\xi,
\end{align}  
where $|\cdot|$ represents the determinant of the matrix. We further provide the first and second order partial derivatives of the collision probability with respect to the centroid of the EV using the Leibniz integral rule, which are given by
\begingroup\makeatletter\def\f@size{8.6}\check@mathfonts
\def\maketag@@@#1{\hbox{\m@th\normalsize\normalfont#1}}
\begin{subequations}\label{eqn:derivatives}
	\begin{align}
	&\frac{\partial \, \mathbb{P}\big(\tilde{P}_k\in \,\measuredangle O_j^k\big)}{\partial x_k} 
	= \int\displaylimits_{\measuredangle O_j^k}  \mathcal{N}\Big(\xi;C_{\rm P}x_k-\mu_j^k,\Sigma_j^k\Big) 
	C^\mathrm{T}_{\rm P}\Big(\xi+\mu_j^k-C_{\rm P}x_k\Big)^\mathrm{T} \Big(\Sigma_j^k\Big)^{-1} \, {\rm d}\xi, \label{eqn:derivatives1} \\
	&\frac{\partial^2 \, \mathbb{P}\big(\tilde{P}_k\in \,\measuredangle O_j^k\big)}{\partial (x_k)^2}
	= \int\displaylimits_{\measuredangle O_j^k}  \mathcal{N}\Big(\xi;C_{\rm P}x_k-\mu_j^k,\Sigma_j^k\Big) 
	C^\mathrm{T}_{\rm P}\Bigg( 
	\Big(\Sigma_j^k\Big)^{-1}\Big(\xi+\mu_j^k-C_{\rm P}x_k\Big) \nonumber \\
	&\Big(\xi+\mu_j^k-C_{\rm P}x_k\Big)^\mathrm{T}\Big(\Sigma_j^k\Big)^{-1} 
	-\Big(\Sigma_j^k\Big)^{-1} 
	\Bigg)C_{\rm P} \, {\rm d} \xi.  \label{eqn:derivatives2}
	\end{align} 
\end{subequations}
\endgroup%
Finally, the obstacle constraints that takes into account the collision probabilities can be given by (\ref{eqn:TV_cons1}) together with the following inequalities
\begin{align} \label{eqn:TV_cons2}
\mathbb{P}\Big(\tilde{P}_k\in \,\measuredangle O_j^k\Big) - \overline{\mathbb{P}}<0, \qquad j = 1,\cdots, M,
\end{align}
where $\overline{\mathbb{P}}>0$ is the maximum allowed collision probability between EV and the TVs.

Equations (\ref{eqn:gaussian})-(\ref{eqn:derivatives}) provide all the necessary results that are required for minimizing the collision probability using the gradient-based optimization methods. Nevertheless, the integrals in (\ref{eqn:gaussian})-(\ref{eqn:derivatives}) are subject to the arbitrary linear constraints and there exist no analytical solutions for them. Thus, one may have to approximate these Gaussian integrals using the numerical sampling methods \cite{gessner2020}, which may be too computationally expensive for real-time implementation. 

In order to address this issue, we, alternatively, propose another approach to handle the uncertainties by using the barrier functions introduced in Section~\ref{sec:barrier}. 
Given the TVs having the uncertain positions, the idea is that we can view the distances between the TVs and the EV as the new random variables and penalize the expected costs that are represented as the barrier functions of the random distances between the EV and the TVs, instead of penalizing their collision probabilities that are difficult to compute. 
To this end, we represent the constraints in (\ref{eqn:TV_cons1}) as the following cost terms,
\begin{align} \label{eqn:random_cost}
b\big(\tilde{P}_k\big) = q_1 \exp \Big(q_2\big(\underline{d} - d^k_j\big(\tilde{P}_k, \mu_j^k\big)\big)\Big), \quad j = 1,\cdots, M,
\end{align}
where $\tilde{P}_k\sim \mathcal{N}(C_{\rm P}x_k-\mu_j^k,\Sigma_j^k)$. One should notice that we have changed the arguments of $d^k_j(\cdot)$ in order to explicitly show the dependency of  (\ref{eqn:random_cost}) on the random variable $\tilde{P}_k$. We then penalize the expected values of the barrier functions in (\ref{eqn:random_cost}), which can be computed using the following equation,
\begin{align} \label{eqn:expect_cost}
\mathbb{E} \, \big[ b(\tilde{P}_k) \big] = \int \mathcal{N}\Big(\xi;C_{\rm P}x_k-\mu_j^k,\Sigma_j^k\Big) b\big(\xi\big)  \, {\rm d}\xi.
\end{align}
Furthermore, we provide the expressions for computing the gradient and Hessian of (\ref{eqn:expect_cost}) as follows,
\begingroup\makeatletter\def\f@size{10}\check@mathfonts
\def\maketag@@@#1{\hbox{\m@th\normalsize\normalfont#1}}
\begin{subequations}\label{eqn:derivatives_expect}
	\begin{align}
	&\frac{\partial \, \mathbb{E} \, \big[ b(\tilde{P}_k) \big]}{\partial x_k} 
	 = \mathbb{E} \, \Big[ b(\tilde{P}_k)C^\mathrm{T}_{\rm P}\Big(\tilde{P}_k+\mu_j^k-C_{\rm P}x_k\Big)^\mathrm{T}\Big(\Sigma_j^k\Big)^{-1} \Big],   \label{eqn:derivatives_expect1}     \\[7pt]
	&\frac{\partial^2 \, \mathbb{E} \, \big[ b(\tilde{P}_k) \big]}{\partial (x_k)^2} 
	 = \mathbb{E} \, \Bigg[ b(\tilde{P}_k)  C^\mathrm{T}_{\rm P}\Bigg( 
	\Big(\Sigma_j^k\Big)^{-1}\Big(\tilde{P}_k+\mu_j^k-C_{\rm P}x_k\Big) \Big(\tilde{P}_k+\mu_j^k \nonumber\\ 
	& -C_{\rm P}x_k\Big)^\mathrm{T}\Big(\Sigma_j^k\Big)^{-1} 
	-\Big(\Sigma_j^k\Big)^{-1} 
	\Bigg)C_{\rm P}   \Bigg].  \label{eqn:derivatives_expect2}
	\end{align} 
\end{subequations}
\endgroup%
The equations in (\ref{eqn:expect_cost})-(\ref{eqn:derivatives_expect2}) can be conveniently evaluated using, for instance, the unscented transform (UT) or the extended Kalman filter (EKF) technique, since we successfully get rid of the linear constraints in the Gaussian integrals (one can see (\ref{eqn:gaussian})-(\ref{eqn:derivatives}) in contrast), which makes it possible for real-time implementation.

\vspace{1 em}
\resizebox{0.95\linewidth}{!}{
	\begin{minipage}[l]{0.55\textwidth}
		\begin{algorithm}[H]
			\caption{ILQR Motion Planning}\label{alg:iLQR}
			\begin{algorithmic}[1]
				\Require{$\bold{x}_0$,  $\bold{u}_0$, $\lambda_0$, $\lambda_{\rm max}$, $s$, $T$, $N$}
				\Ensure{$\bold{x}^*$, $\bold{u}^*$}
				\vspace{2 mm}
				\State $\lambda\leftarrow \lambda_0$;
				\State $\bold{x}^-\leftarrow \bold{x}_0, \bold{u}^-\leftarrow\bold{u}_0$;
				\State Convert constraints (\ref{eqn:u_cons})-(\ref{eqn:expect_cost}) to costs using barrier function (\ref{eqn:exp});
				\State $J^-$, $\big\{(f_x)_k,(f_u)_k,(\ell_x)_k, (\ell_u)_k, (\ell_{xx})_k, (\ell_{xu})_k, (\ell_{uu})_k\big\}_{k=0,\cdots,T}$ $\gets$ forward pass using control sequence $\bold{u}_0$;
				\State Convergence $\gets$ $\textbf{False}$;
				\For{$i\gets0$ to $N$} 
				\State $\big\{H_k,G_k \big\}_{k=0,\cdots,T}\gets$ backward pass following (\ref{eqn:value})-(\ref{eqn:value_x}) and (\ref{eqn:inv_P});
				\vspace{-0.9 em}
				\State $J^+$, $\bold{x}^+$, $\bold{u}^+$, $\big\{(f_x)_k,(f_u)_k,(\ell_x)_k, (\ell_u)_k, (\ell_{xx})_k, (\ell_{xu})_k, (\ell_{uu})_k\big\}_{k=0,\cdots,T}$ $\gets$ forward pass using $\big\{H_k,G_k \big\}_{k=0,\cdots,T}$;
				\If{$J^+<J^-$} 
				\State $\lambda\gets \lambda/s$;
				\State $J^-\gets J^+$;
				\State $\bold{x}^-\gets\bold{x}^+$, $\bold{u}^-\gets\bold{u}^+$;
				\If{Convergence}
				\State $\textbf{break}$;
				\EndIf
				\Else
				\State $\lambda\gets s\lambda$;
				\If{$\lambda>\lambda_{\rm max}$}
				\State $\textbf{break}$;
				\EndIf
				\EndIf
				\EndFor
				\vspace{2 mm}
				\State \Return $\bold{x}^*\gets \bold{x}^-$, $\bold{u}^*\gets \bold{u}^-$;
			\end{algorithmic}
		\end{algorithm}
\end{minipage} }%


\subsection{Levenberg-Marquadt} \label{sec:Implementation}
It is worth mentioning that the quadratic approximations in (\ref {eqn:P_quadratic}) and (\ref{eqn:b_quadratic}) cannot guarantee the Hessian matrices (i.e., $P_{xx}$, $P_{uu}$) to be positive-definite all the time, which depends on the nonlinear dynamics of the system in (\ref{eqn:eom1})-(\ref{eqn:eom4}) and the nonlinear cost functions in (\ref{eqn:control})-(\ref{eqn:derivatives_expect}) for a specific motion planning problem. Nevertheless, the optimal control law computed by (\ref {eqn:u}) requires the positive definiteness of $P_{uu}$ at each time step during an iLQR iteration. 
To fix this issue, one could use a Levenberg-Marquadt method \cite{todorov2005}, an adaptive shift scheme \cite{liao1992}, or simply fix the Hessian to be identity. 

This paper uses the well-known Levenberg-Marquadt method to fix the Hessian and search for the optimal trajectory iteratively by appropriately adjusting the so called damping factor. The inverse matrix of the Hessian $P_{uu}$ is computed as follows: 
1) We first decompose the Hessian $P_{uu} = U\Lambda U^{\mathrm{T}}$ since $P_{uu}$ is symmetric, where $U^{\mathrm{T}}U=\boldsymbol{I}$ are the normalized eigenvectors and $\Lambda$ is the diagonal matrix of the eigenvalues.
2) Next, we set all the negative eigenvalues in $\Lambda$ to 0;
3) The inverse matrix of $P_{uu}$ in (\ref{eqn:value_x}) is given by
\begin{align} \label{eqn:inv_P}
P^{-1}_{uu} = U\big(\Lambda+\lambda \boldsymbol{I} \big)^{-1}U^\mathrm{T},
\end{align}
where $\lambda>0$ is the damping factor. We then summarize the iLQR motion planning algorithm in 
Algorithm~\ref{alg:iLQR} (see Table~\ref{tab:parameters} for the unknown parameters).

\section{Results} \label{sec:result}
In this section we implement Algorithm~\ref{alg:iLQR} for emergent collision avoidance in different scenarios in simulation. Furthermore, we validate the motion planning algorithm in multiple real-time tasks using both a simulator and a level-3 autonomous driving test platform developed by the Tencent Autonomous Driving team, which, for short, are referred to as TadSim and TadAuto, respectively. 

Since the existing work in the literature has provided sufficient comparison results between iLQR and the standard baseline motion planning algorithms (i.e., SQP) \cite{chen2019auto}, which shows the great advantage of iLQR in the computation efficiency by saving 98\% running time than the SQP per iteration, this paper does not repeat such comparison work as this is already done well. We, instead, concentrate on validating the iLQR algorithm in the more challenging and more complicated tasks that were not investigated enough in the existing work, such as the real-time emergent collision avoidance in the real-world dynamically-changing and uncertain traffic.

\subsection{Simulation}
First, we simulate the behaviors of several environmental TVs in Python 3.7 and implement Algorithm~\ref{alg:iLQR} with the MDR formulation for collision avoidance. The planning horizon for the EV at each time step is set to $\mathcal{T}=5$ seconds, during which the predicted trajectories of the surrounding TVs are assumed to be exactly given. The time interval for the motion planning of the EV is ${\rm dt}=0.25$ seconds.  
Other related design parameters for the iLQR algorithm are summarized in Table~\ref{tab:parameters}.
\begin{table}[H]
	\small
	\setlength{\belowcaptionskip}{1pt}
	\centering
	\caption{iLQR design parameters \& simulation condition.}  \label{tab:parameters}
	\scalebox{0.67}{
		\begin{tabular}{|c|c|c|c|c|c|}
			\hline
			$\mathcal{T}$ [sec] & 5.0 & planning horizon & $N$ [-] & 20 & max iteration Num.\\
			\hline
			${\rm dt}$ [sec] & 0.25 & time interval & $s$ [-] & 5e2 & scale \\
			\hline
			$\lambda_0$ [-] & 1.0 & initial damping & $\lambda_{\rm max}$ [-] & 1e10 & max damping \\
			\hline			
			$a_{\rm max}$ [m/sec$^2$] & 2.0 & max acceleration & $a_{\rm min}$ [m/sec$^2$] & -4.0 & min acceleration \\
			\hline
			$r_{\rm max}$ [rad/sec] & 0.25 & max yaw rate & $r_{\rm min}$ [rad/sec] & -0.25 & min yaw rate \\			
			\hline	
			$\underline{d}$ [m] & 1.0 & min EV/TV distance & $\overline{\mathbb{P}}$ [-] & 0.1 & max collision risk \\			
			\hline												
	\end{tabular} }	
	\vspace{0.2 em}

	\scalebox{0.785}{
	\begin{tabular}{|c|c|c|c|c|c|c|c|c|c|c|c|}
		\hline
		$w_{\rm a}$& 1e3 & $w_{\rm r}$ & 1e5 & $w_{\rm p^x}$ & 1&  $w_{\rm p^y}$ & 1&$w_{\rm v}$ & 1e4 &$w_{\rm \psi}$& 1e4\\
		\hline
		$w_{\rm p^{\rm ref}}$ & 1e5 &  $w_{\rm v^{\rm ref}}$ & 1e3&  $w^{\rm f}_{\rm \psi}$& 1e4 &$w^{\rm f}_{\rm v}$ & 1e3 & $q_1$ & 1e2 & $q_2$ & 10 \\	
		\hline													
	\end{tabular} }	
	\vspace{-0.1 em}	
	
	\scalebox{0.61}{
	\begin{tabular}{|c|c|c|c|c|c|c|}

		\hline
		       & speed [m/s] & lat. distance [m] & long. distance [m] & length [m] & width [m] & task\\
		\hline
		EV     & 20.0 & - & - & 5.0 & 2.0&  lane-keeping \\
		\hline
		${\rm TV}_1$ & 10.0 & -2.0 & 15.0 &  5.0 & 2.0 & switch to left \\
		\hline
		${\rm TV}_2$ & 10.0 & 4.0  & 0.0 & 5.0 & 2.0 & lane-keeping \\		
		\hline	
		${\rm TV}_3$ & 12.0 & -4.0 & -10.0 & 5.0 & 2.0 & lane-keeping \\	
		\hline															
\end{tabular} }	
\end{table}

\begin{figure}[!htbp]
	\centering
	\includegraphics[width=0.47\textwidth,height=0.32\textwidth]{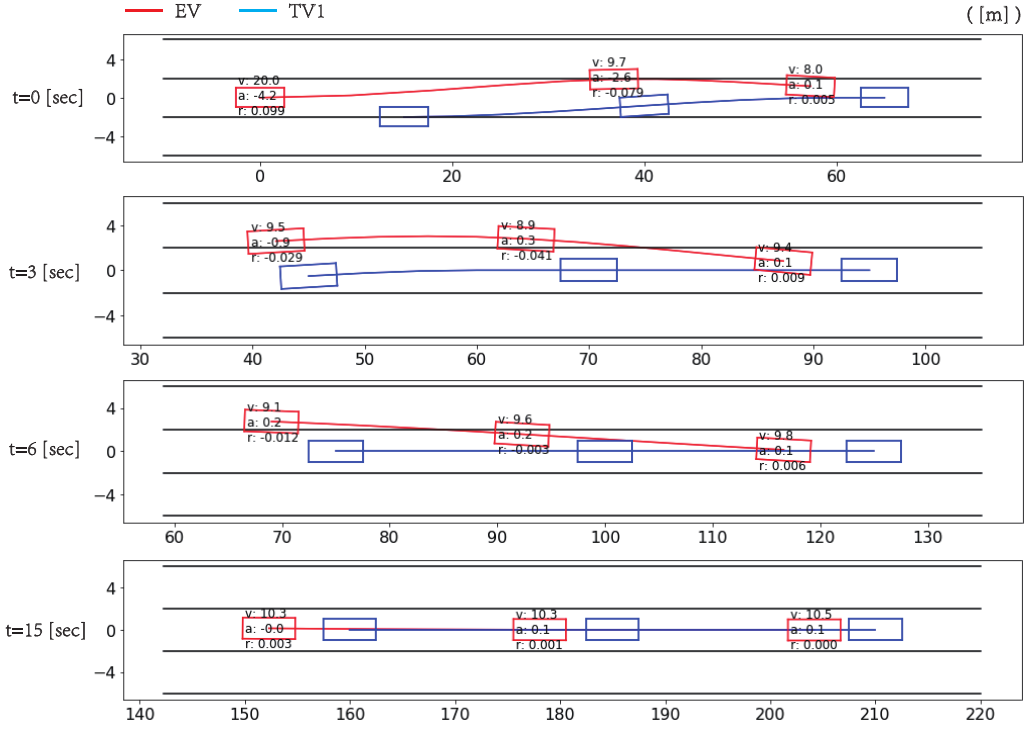}
	\caption{Collision avoidance in a 3-lane scenario with single TV.}
	\label{fig:6_result1}
\end{figure}

We first consider a typical emergent cut-in scenario where a low speed TV suddenly switches to the lane of the EV using about 2 seconds from a neighbour lane. One can find the simulation conditions for the EV and ${\rm TV}_1$ from Table~\ref{tab:parameters} in this case. Since we assume the minimum acceleration of the EV to be $a_{\rm min}=-4.0$ [m/sec$^2$], the minimum safe distance between EV and ${\rm TV}_1$ should be at least $(V_{\rm EV}-V_{{\rm TV}_1})^2/2|a_{\rm min}|+(l_{\rm EV}+l_{{\rm TV}_1})/2 = 17.5$ [m] (larger than 15 [m]), where $V_{\rm EV}$, $V_{{\rm TV}_1}$, $l_{\rm EV}$ and $l_{{\rm TV}_1}$ are the initial speeds and the vehicle lengths of the EV and ${\rm TV}_1$, respectively. This result indicates that the EV cannot avoid collision with ${\rm TV}_1$ by braking merely without proper lateral motion.
Fig.~\ref{fig:6_result1} plots the planned trajectory of the EV and the predicted trajectory of ${\rm TV}_1$ over $\mathcal{T}$ at different time instants, where the symbols $v$, $a$, and $r$ denote the velocity [m/s], the acceleration [m/$s^2$] and the yaw rate [rad/s] of the EV, respectively. The result shows that the EV is able to avoid the collision with ${\rm TV}_1$ by temporally taking up the left neighbour lane which is available when ${\rm TV}_1$ is changing the lane. 

\begin{figure}[!htbp]
	\centering
	\includegraphics[width=0.47\textwidth,height=0.31\textwidth]{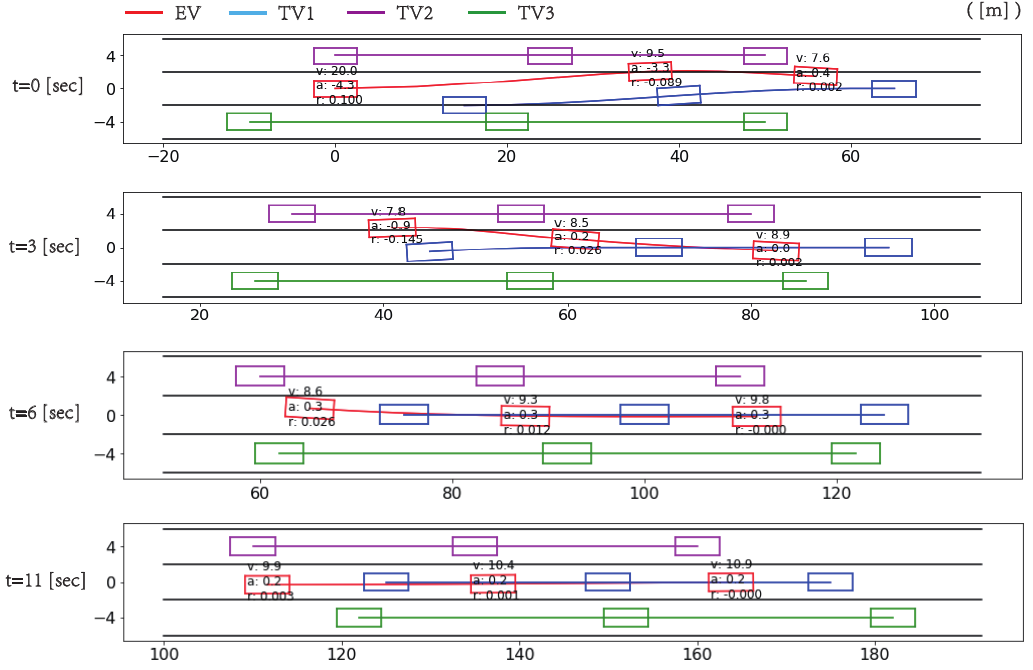}
	\caption{Collision avoidance in a 3-lane scenario with multiple TVs.}
	\label{fig:6_result2}
\end{figure}

Next, we make the collision avoidance task more challenging by including more TVs into the driving scenario (see ${\rm TV}_2$ and ${\rm TV}_3$ in Table~\ref{tab:parameters}), such that both the left lane and the right lane of the EV will be occupied when the cut-in behavior of ${\rm TV}_1$ takes place. The new simulation trajectories are plotted in Fig.~\ref{fig:6_result2}. 

The result in Fig.~\ref{fig:6_result2} shows that the EV is able to avoid collision with both ${\rm TV}_1$ and ${\rm TV}_2$ by more reasonably using the space of its current lane and the space of the left neighbour lane. By comparing the results in Fig.~\ref{fig:6_result1} and Fig.~\ref{fig:6_result2} one can see that in the first case half of the left lane was occupied by the EV for collision avoidance, while in the second case only a quarter of the left lane or so was occupied such that the EV is able to keep certain safe distance from ${\rm TV}_2$ at the same time.

\begin{figure}[!htbp]
	\centering
	\includegraphics[width=0.465\textwidth,height=0.32\textwidth]{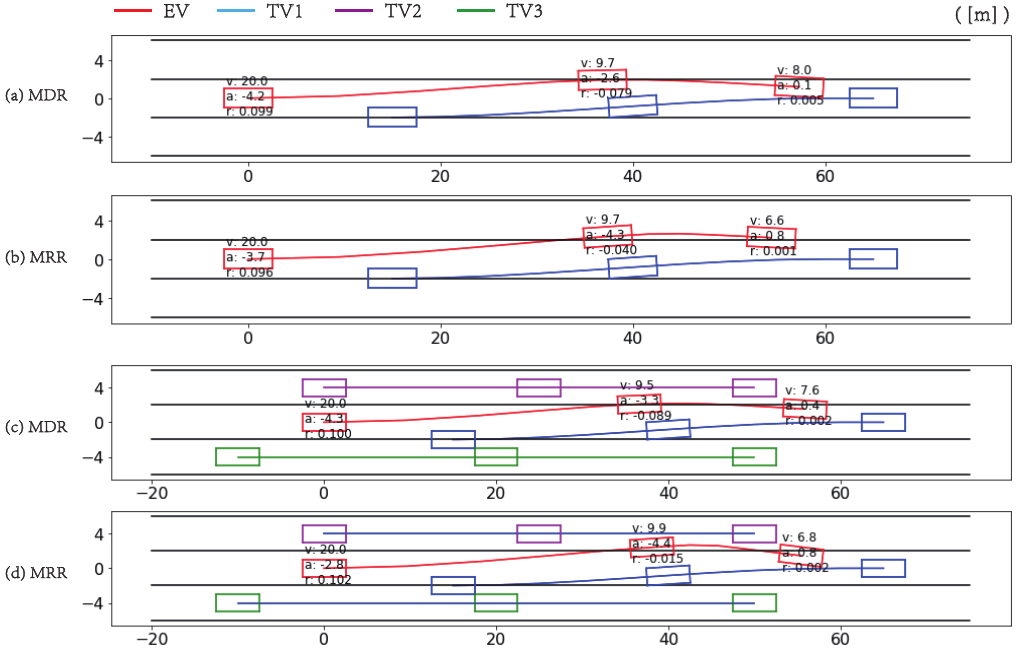}
	\caption{Comparison between MDR \& MRR in both cases.}
	\label{fig:6_result3}
\end{figure}

Furthermore, we repeat the two tasks above in Fig.~\ref{fig:6_result1} and Fig.~\ref{fig:6_result2} by implementing Algorithm~\ref{alg:iLQR} with the MRR formulation, respectively, while assuming each point of the predicted trajectories of the TVs to be suffering from the uniform covariance 
$\Sigma_j^k = 0.25\times \boldsymbol{I}_{2}$.
The simulation results are shown in Fig.~\ref{fig:6_result3}.
By comparing Fig.~\ref{fig:6_result3} (a) and Fig.~\ref{fig:6_result3} (b) one sees that the planned trajectory for the EV using the MRR formulation shows more bending in the middle of the trajectory, since we assume the position of ${\rm TV}_1$ is not exactly predicted and a larger distance between the two vehicles is required for better the safety. By comparing Fig.~\ref{fig:6_result3} (b) and Fig.~\ref{fig:6_result3} (d) one sees that the planned trajectory for the EV shows more bending at the end of the trajectory in order to avoid the collision with  ${\rm TV}_2$ on the left, whose position is also not exactly predicted by assumption. This result shows the effectiveness of the MRR method to handle the  environmental obstacles' uncertainties for collision avoidance. Nonetheless, the EV has to keep a larger distance from the surrounding TVs in the uncertain traffic environment, which looks quite conservative and the smoothness of the EV's trajectory also gets worse. Hence, it is still important to enhance the environmental perception ability of the self-driving vehicles and further improve the trajectory prediction quality for the TVs to minimize the effect of the uncertainties for the better self-driving performance.

\subsection{Real-Time Implementation}
\begin{figure}[!htbp]
	\centering
	\includegraphics[width=0.49\textwidth,height=0.16\textwidth]{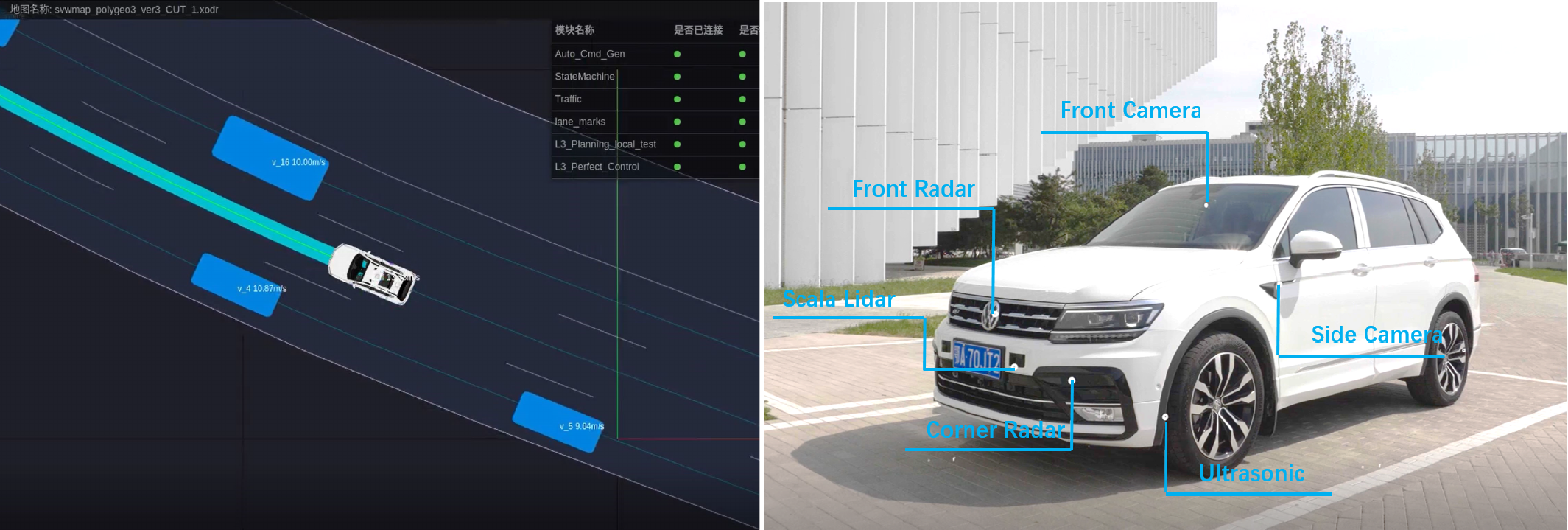}
	\caption{Simulation interface of TadSim (left) and TadAuto (right).}
	\label{fig:7_tadsim}
\end{figure}

Next, we implement the motion planning algorithm in real-time using TadSim and TadAuto, respectively. TadSim is a real-time traffic simulation platform that is built for developing the self-driving technology,  where the self-driving vehicle is modeled with 27 degrees of freedom and the traffic vehicles are mainly controlled by the intelligent driver model \cite{treiber2000}. TadSim is especially suitable for generating the diverse high-fidelity vehicle maneuvers in all kinds of driving scenarios. Fig.~\ref{fig:7_tadsim} shows the interface of TadSim, which includes an EV in the middle and multiple TVs driving on a high-definition map. We manually design more than a hundred of testing cases by mimicking the various cut-in maneuvers of the TVs using TadSim, and implement the iLQR motion planing algorithm for collision avoidance in all testing cases. The algorithm is coded in C++ 11 to obtain the better implementation speed. 

\begin{table}[H]
	\small
	\setlength{\belowcaptionskip}{1pt}
	\centering
	\caption{Simulation result with/without iLQR.}  \label{tab:result}
	\scalebox{0.8}{
		\begin{tabular}{|c|c|c|c|}
			\hline
			& Ave. acceleration [m/$s^2$] & Ave. jerk [m/$s^3$] & Accident rate \\
			\hline
			Braking-only & -1.32 & 1.25 & 14/121 \\
			\hline
			iLQR         & -0.25 & 0.84 & 0/121 \\
			\hline
			Improvement  & 81.1\% & 32.8\%  & -  \\		
			\hline																
	\end{tabular} }	
\end{table}

Table~\ref{tab:result} summarizes the simulation results.  The item ``braking-only'' corresponds to the experiments where only the longitudinal control (i.e., intelligent driver model) is applied for collision avoidance during lane-keeping.  One sees that the iLQR motion planning algorithm is able to safely pass all the testing cases without any collision, while providing the better driving comfort by obviously improving the average acceleration and the average jerk.  

Furthermore, we implement the iLQR motion planing algorithm on the TadAuto platform in the real-world traffic environment, where TadAuto is a level-3 autonomous vehicle which is mainly equipped with five radars and a front-view camera. The real-world field experiments show similar results with the simulations using TadSim (see Fig.~\ref{fig:7_tadsim}).  We demonstrate the collision avoidance behavior of the self-driving vehicle using the video which can be found  via the link below \footnote{The videos are available on the youtube channel: \url{https://youtu.be/BhR98UpTDEg}}. We do not provide more results due to the limit of space.



\section{Conclusion} \label{sec:Conclusion}

We propose a new iLQR motion planning algorithm for developing the self-driving technology, which concentrates on handling the uncertain behaviors of the TVs in the traffic for emergent collision avoidance during lane-keeping. This approach can be easily extended for more driving scenarios such as lane-changing and on/off ramp merging by adjusting a little bit the design of the cost functions.

We use the so-called ``collision polygon'' to determine the minimum distance between the EV and each TV of interest in the traffic, and introduce two different methods for designing the constraints of the motion planning problem to handle the uncertain behaviors of the surrounding TVs. The iLQR motion planning algorithm is further validated using both the simulation results and the real-word experiments for collision avoidance in traffic. Promising results are observed which show both the implementation efficiency and the effectiveness of the iLQR motion planning algorithm in multiple manually designed driving experiments, while improving obviously the driving comfort simultaneously.

Future work will focus on improving the work to incorporate special obstacles such as the pedestrians that may need extra attention for motion planning. Also more experiments and test cases considering different levels of the uncertainties in the state estimations for the obstacles will be designed for further possible improvement of the driving safety.  

\section*{ACKNOWLEDGMENT}
The authors would like to thank the Tencent autonomous driving team for providing the necessary assistance for conducting the experiments.

\bibliographystyle{IEEEtran}
\bibliography{changxi}

\begin{thebibliography}{10}
\providecommand{\url}[1]{#1}
\csname url@rmstyle\endcsname
\providecommand{\newblock}{\relax}
\providecommand{\bibinfo}[2]{#2}
\providecommand\BIBentrySTDinterwordspacing{\spaceskip=0pt\relax}
\providecommand\BIBentryALTinterwordstretchfactor{4}
\providecommand\BIBentryALTinterwordspacing{\spaceskip=\fontdimen2\font plus
\BIBentryALTinterwordstretchfactor\fontdimen3\font minus
  \fontdimen4\font\relax}
\providecommand\BIBforeignlanguage[2]{{%
\expandafter\ifx\csname l@#1\endcsname\relax
\typeout{** WARNING: IEEEtran.bst: No hyphenation pattern has been}%
\typeout{** loaded for the language `#1'. Using the pattern for}%
\typeout{** the default language instead.}%
\else
\language=\csname l@#1\endcsname
\fi
#2}}

\bibitem{Brechtel2011}
S.~Brechtel, T.~Gindele, and R.~Dillmann, ``Probabilistic {MDP}-behavior
  planning for cars,'' in \emph{14th International IEEE Conference on
  Intelligent Transportation Systems}, Washington, DC, October 5--7 2011, pp.
  1537--1542.

\bibitem{Kuwata2009}
Y.~Kuwata, J.~Teo, G.~Fiore, S.~Karaman, E.~Frazzoli, and J.~P. How,
  ``Real-time motion planning with applications to autonomous urban driving,''
  \emph{IEEE Transactions on Control Systems Technology}, vol.~17, no.~5, pp.
  1105--1118, 2009.

\bibitem{Karaman2011}
S.~Karaman and E.~Frazzoli, ``Sampling-based algorithms for optimal motion
  planning,'' \emph{The International Journal of Robotics Research}, vol.~30,
  no.~7, pp. 846--894, 2011.

\bibitem{velenis2005optimal}
E.~Velenis and P.~Tsiotras, ``Optimal velocity profile generation for given
  acceleration limits: theoretical analysis,'' in \emph{Proceedings of the IEEE
  American Control Conference}, vol.~2, 2005, pp. 1478--1483.

\bibitem{you2019tcst}
C.~{You} and P.~{Tsiotras}, ``High-speed cornering for autonomous off-road
  rally racing,'' \emph{IEEE Transactions on Control Systems Technology}, pp.
  1--17, 2019.

\bibitem{Jaillet2010}
L.~Jaillet, J.~Cort{\'e}s, and T.~Sim{\'e}on, ``Sampling-based path planning on
  configuration-space costmaps,'' \emph{IEEE Transactions on Robotics},
  vol.~26, no.~4, pp. 635--646, 2010.

\bibitem{Garcia2013}
M.~Garcia, A.~Viguria, and A.~Ollero, ``Dynamic graph-search algorithm for
  global path planning in presence of hazardous weather,'' \emph{Journal of
  Intelligent \& Robotic Systems}, vol.~69, no. 1-4, pp. 285--295, 2013.

\bibitem{Cimurs2017}
R.~Cimurs, J.~Hwang, and I.~H. Suh, ``{B\'{e}zier} curve-based smoothing for
  path planner with curvature constraint,'' in \emph{IEEE International
  Conference on Robotic Computing}, Taichung, Taiwan, April 10--12 2017, pp.
  241--248.

\bibitem{Choi2008}
J.-w. Choi, R.~Curry, and G.~Elkaim, ``Path planning based on {B}{\'e}zier
  curve for autonomous ground vehicles,'' in \emph{World Congress on
  Engineering and Computer Science}, San Francisco, CA, October 22--24 2008,
  pp. 158--166.

\bibitem{Choi2010}
J.-w. Choi, R.~E. Curry, and G.~H. Elkaim, ``Continuous curvature path
  generation based on {B}{\'e}zier curves for autonomous vehicles.''
  \emph{International Journal of Applied Mathematics}, vol.~40, no.~2, pp.
  91--101, 2010.

\bibitem{Shim2012}
T.~Shim, G.~Adireddy, and H.~Yuan, ``Autonomous vehicle collision avoidance
  system using path planning and model-predictive-control-based active front
  steering and wheel torque control,'' \emph{Proceedings of the Institution of
  Mechanical Engineers, Part D: Journal of automobile engineering}, vol. 226,
  no.~6, pp. 767--778, 2012.

\bibitem{wang2017real}
J.~Wang, M.~A. Garratt, and S.~G. Anavatti, ``Real-time path planning algorithm
  for autonomous vehicles in unknown environments,'' \emph{International
  Journal of Mechatronics and Automation}, vol.~6, no.~1, pp. 1--9, 2017.

\bibitem{Chee1994}
W.~Chee and M.~Tomizuka, ``Vehicle lane change maneuver in automated highway
  systems,'' Institute Of Transportation Studies, University of California,
  Berkeley, Tech. Rep. UCB-ITS-PRR-94-22, 1994.

\bibitem{Fraichard2004}
T.~Fraichard and A.~Scheuer, ``From {Reeds} and {Shepp's} to
  continuous-curvature paths,'' \emph{IEEE Transactions on Robotics}, vol.~20,
  no.~6, pp. 1025--1035, 2004.

\bibitem{chen2013lane}
J.~Chen, P.~Zhao, T.~Mei, and H.~Liang, ``Lane change path planning based on
  piecewise bezier curve for autonomous vehicle,'' in \emph{Proceedings of 2013
  IEEE International Conference on Vehicular Electronics and Safety}, 2013, pp.
  17--22.

\bibitem{Korzeniowski2016}
D.~Korzeniowski and G.~{\'S}laski, ``Method of planning a reference trajectory
  of a single lane change manoeuver with {B\'{e}zier} curve,'' in \emph{IOP
  Conference Series: Materials Science and Engineering}, vol. 148, no.~1.\hskip
  1em plus 0.5em minus 0.4em\relax IOP Publishing, 2016, p. 012012.

\bibitem{you2019autonomous}
C.~You, J.~Lu, D.~Filev, and P.~Tsiotras, ``Autonomous planning and control for
  intelligent vehicles in traffic,'' \emph{IEEE Transactions on Intelligent
  Transportation Systems}, vol.~21, no.~6, pp. 2339--2349, 2020.

\bibitem{kushleyev2009time}
A.~Kushleyev and M.~Likhachev, ``Time-bounded lattice for efficient planning in
  dynamic environments,'' in \emph{IEEE International Conference on Robotics
  and Automation}, Kobe, Japan, 2009, pp. 1662--1668.

\bibitem{pivtoraiko2009diff}
M.~Pivtoraiko, R.~A. Knepper, and A.~Kelly, ``Differentially constrained mobile
  robot motion planning in state lattices,'' \emph{Journal of Field Robotics},
  vol.~26, no.~3, pp. 308--333, 2009.

\bibitem{ziegler2009space}
J.~Ziegler and C.~Stiller, ``Spatiotemporal state lattices for fast trajectory
  planning in dynamic on-road driving scenarios,'' in \emph{IEEE/RSJ
  International Conference on Intelligent Robots and Systems}, 2009, pp.
  1879--1884.

\bibitem{mcnaughton2011motion}
M.~McNaughton, C.~Urmson, J.~M. Dolan, and J.-W. Lee, ``Motion planning for
  autonomous driving with a conformal spatiotemporal lattice,'' in \emph{IEEE
  International Conference on Robotics and Automation}, Shanghai, China, 2011,
  pp. 4889--4895.

\bibitem{chandra2017safe}
R.~Chandra, Y.~Selvaraj, M.~Br{\"a}nnstr{\"o}m, R.~Kianfar, and N.~Murgovski,
  ``Safe autonomous lane changes in dense traffic,'' in \emph{IEEE 20th
  International Conference on Intelligent Transportation Systems}, 2017, pp.
  1--6.

\bibitem{obayashi2018real}
M.~Obayashi and G.~Takano, ``Real-time autonomous car motion planning using
  nmpc with approximated problem considering traffic environment,''
  \emph{IFAC-PapersOnLine}, vol.~51, no.~20, pp. 279--286, 2018.

\bibitem{chen2019auto}
J.~Chen, W.~Zhan, and M.~Tomizuka, ``Autonomous driving motion planning with
  constrained iterative {LQR},'' \emph{IEEE Transactions on Intelligent
  Vehicles}, vol.~4, no.~2, pp. 244--254, 2019.

\bibitem{zhao2018}
Y.~Zhao, Y.~Wang, M.~Zhou, and J.~Wu, ``Energy-optimal collision-free motion
  planning for multiaxis motion systems: An alternating quadratic programming
  approach,'' \emph{IEEE Transactions on Automation Science and Engineering},
  vol.~16, no.~1, pp. 327--338, 2018.

\bibitem{pan2020safe}
Y.~Pan, Q.~Lin, H.~Shah, and J.~M. Dolan, ``Safe planning for self-driving via
  adaptive constrained {ILQR},'' \emph{arXiv preprint arXiv:2003.02757}, 2020.

\bibitem{hart1968formal}
P.~E. Hart, N.~J. Nilsson, and B.~Raphael, ``A formal basis for the heuristic
  determination of minimum cost paths,'' \emph{IEEE transactions on Systems
  Science and Cybernetics}, vol.~4, no.~2, pp. 100--107, 1968.

\bibitem{koenig2002d}
S.~Koenig and M.~Likhachev, ``D* lite,'' \emph{Aaai/iaai}, vol.~15, 2002.

\bibitem{gonzalez2015review}
D.~Gonz{\'a}lez, J.~P{\'e}rez, V.~Milan{\'e}s, and F.~Nashashibi, ``A review of
  motion planning techniques for automated vehicles,'' \emph{IEEE Transactions
  on Intelligent Transportation Systems}, vol.~17, no.~4, pp. 1135--1145, 2015.

\bibitem{mayne1973diff}
D.~Q. Mayne, ``Differential dynamic programming--a unified approach to the
  optimization of dynamic systems,'' in \emph{Control and Dynamic
  Systems}.\hskip 1em plus 0.5em minus 0.4em\relax Elsevier, 1973, vol.~10, pp.
  179--254.

\bibitem{chen2017}
J.~Chen, W.~Zhan, and M.~Tomizuka, ``Constrained iterative lqr for on-road
  autonomous driving motion planning,'' in \emph{IEEE 20th International
  Conference on Intelligent Transportation Systems}, 2017, pp. 1--7.

\bibitem{gill2005}
P.~E. Gill, W.~Murray, and M.~A. Saunders, ``{SNOPT}: An {SQP} algorithm for
  large-scale constrained optimization,'' \emph{SIAM review}, vol.~47, no.~1,
  pp. 99--131, 2005.

\bibitem{biegler2009}
L.~T. Biegler and V.~M. Zavala, ``Large-scale nonlinear programming using
  {IPOPT}: An integrating framework for enterprise-wide dynamic optimization,''
  \emph{Computers \& Chemical Engineering}, vol.~33, no.~3, pp. 575--582, 2009.

\bibitem{todorov2005}
E.~Todorov and W.~Li, ``A generalized iterative {LQG} method for
  locally-optimal feedback control of constrained nonlinear stochastic
  systems,'' in \emph{Proceedings of the IEEE American Control Conference},
  2005, pp. 300--306.

\bibitem{philipp2019}
A.~Philipp and D.~Goehring, ``Analytic collision risk calculation for
  autonomous vehicle navigation,'' in \emph{International Conference on
  Robotics and Automation}, 2019, pp. 1744--1750.

\bibitem{gessner2020}
A.~Gessner, O.~Kanjilal, and P.~Hennig, ``Integrals over {G}aussians under
  linear domain constraints,'' in \emph{International Conference on Artificial
  Intelligence and Statistics}.\hskip 1em plus 0.5em minus 0.4em\relax PMLR,
  2020, pp. 2764--2774.

\bibitem{liao1992}
L.-z. Liao and C.~A. Shoemaker, ``Advantages of differential dynamic
  programming over {N}ewton's method for discrete-time optimal control
  problems,'' Cornell University, Tech. Rep., 1992.

\bibitem{treiber2000}
M.~Treiber, A.~Hennecke, and D.~Helbing, ``Congested traffic states in
  empirical observations and microscopic simulations,'' \emph{Physical review
  E}, vol.~62, no.~2, p. 1805, 2000.

\end{thebibliography}

\vspace*{-15 mm}
\begin{IEEEbiography}
	[{\includegraphics[width=1in,height=1.25in,clip,keepaspectratio]{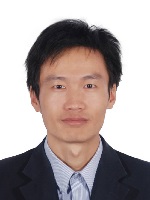}}]{Changxi You} is a senior researcher in Autonomous Driving Laboratory at Tencent Technology Company.
He received his B.S. and M.S. degrees from the Department of Automotive Engineering, Tsinghua University of China, and a second M.S. degree from the Department of Automotive Engineering, RWTH-Aachen University of Germany. He received his Ph.D. degree under the supervision of Prof. Panagiotis Tsiotras from the School of Aerospace Engineering, Georgia Institute of Technology. His current research interests are in motion planning \& control in congested traffic, collision avoidance in uncertain environment, and intelligent coordination and control for large-scale traffic signals.
\end{IEEEbiography}

\end{document}